# Machine learning-based system reliability analysis with Gaussian Process Regression

Lisang Zhou[1], Ziqian Luo[2*], Xueting Pan[3]

[1]Bazaarvoice Inc., Austin, 78759, TX, United States, Email: lzhou@berkeley.edu
[2]Oracle, Seattle, 98101, WA, United States, Corresponding author, Email: luoziqian98@gmail.com
[3]Oracle, Seattle, 98101, WA, United States, Email: xtpan8800@gmail.com

## ABSTRACT

Machine learning-based reliability analysis methods have shown great advancements for their computational efficiency and accuracy. Recently, many efficient learning strategies have been proposed to enhance the computational performance. However, few of them explores the theoretical optimal learning strategy. In this article, we propose several theorems that facilitates such exploration. Specifically, cases that considering and neglecting the correlations among the candidate design samples are well elaborated. Moreover, we prove that the well-known $U$ learning function can be reformulated to the optimal learning function for the case neglecting the Kriging correlation. In addition, the theoretical optimal learning strategy for sequential multiple training samples enrichment is also mathematically explored through the Bayesian estimate with the corresponding lost functions. Simulation results show that the optimal learning strategy considering the Kriging correlation works better than that neglecting the Kriging correlation and other state-of-the art learning functions from the literatures in terms of the reduction of number of evaluations of performance function. However, the implementation needs to investigate very large computational resource.

**Key words**: *Reliability analysis; Risk analysis; Surrogate models; Kriging; Gaussian Process Regression; Active Learning;*

## 1. Introduction

Probabilities of occurrence of risky events are critically decisive for engineers and researchers to quantify the risk of planed operations and conductions. These probabilities are typically characterized by conducting reliability analysis, of which the target is to estimate the probability of failure, denoted as $P_f$. In reliability analysis, $P_f$ can be calculated as:

$$P_f = \int_{g(\boldsymbol{x})\leq 0} \rho(\boldsymbol{x}) d\boldsymbol{x} = \int_{\Omega} I_g(\boldsymbol{x})\rho(\boldsymbol{x}) d\boldsymbol{x}\,, \tag{1}$$

where $\boldsymbol{x}$ is the vector of random variables, $g(\boldsymbol{x})$ is the so-called performance or limit state function, $\rho(\boldsymbol{x})$ is the joint probability density function (PDF) of $\boldsymbol{x}$ and $I_g(\boldsymbol{x})$ is a failure indicator function. Note that $I_g(\boldsymbol{x}) = 1$ when $g(\boldsymbol{x}) \leq 0$, and $I_g(\boldsymbol{x}) = 0$ when $g(\boldsymbol{x}) > 0$. Modern reliability analysis methodologies aim to estimate $P_f$ with as less as prossible number of evluations to the computationally demanding numerical models. Those well-developed techniques primarily include simulation-based sampling techniques (e.g., the crude Monte-Carlo simulation (MCS) [1], [2], importance sampling (IS) [3], and subset simulation (SS) [4]) and approximation-based approaches (e.g. first- or second-order reliability analysis methods) [5], [6]), which estimate $P_f$ by searching for the most probable point in the probabilistic space. Despite the first group can produce desirable estimate of $P_f$, it is computationally expensive. On the contrary, approximation-based methods are often computationally fast, they lack the accuracy for problems with non-linear responses near the limit state. To address aforementioned limitations, surrogate model-based reliability analysis has emerged and shown advancement. Those surrogate models are mainly response surfaces [7], [8], artificial neutral networks [9], [10], [11], support vector machines [12], [13], polynomial chaos expansions [14], [15] and the Gaussian Process Regression or Kriging model [16], [17]. Among these techniques, the Kriging model has been developing extremely fast out of its inherent advantages [18], [19], [20], [21].

The Kriging model or Gaussian process regression is a Bayesian regression model that combines interpolation and regression. Different from other deterministic regression model, outputs from the Kriging follow Gaussian distribution with corresponding mean and variance. Wang et al [22] derives the formulations for Kriging model with correlated responses. This stochastic property of Kriging model has been taken advantage to construct the active learning-based reliability analysis methods, where the training sample set is adaptively enriched to construct a well-trained surrogate model. The well-trained surrogate model can subsequently substitute the sophisticated model for the process of Monte Carlo simulations. Two representative active learning-based reliability analysis methods have been widely accepted: Efficient Global Reliability Analysis (EGRA) proposed by Bichon et al. [17] and Adaptive Kriging-based Monte Carlo simulation (AK-MCS) proposed by Echard et al. [16]. Different from EGRA, candidate design samples in AK-MCS is adaptively increased to ensure that the coefficient of variation of the failure probability is smaller than a prescribed threshold.

It is known that the learning strategy for properly selecting training points plays a decisive role in Kriging-based reliability analysis methods. Aside from EFF and U learning functions, a lot of creative developments and variations for these two methods have also been proposed by researchers recently to enhance the methodologies. For the variations of learning functions, *H* learning function takes the advantage of proposed by Lv et al. [23], least improvement function (*LIF*) proposed by Sun et al. [24] and $\boldsymbol{\psi}_d$ and $\boldsymbol{\psi}_\sigma$ proposed by Xiao et al. [25] have shown great advantages in selecting next best training point. For details, *H* learning function aims to seek the training point in the vicinity of the limit state, which follows the same principle as *EFF* but in a way of information entropy theory. Moreover, *LIF* takes advantage of probability density of each point to highlight the training points with large probability densities. They pick the best training point that deviates from the existing training points to avoid the ill-conditioning problem, and is close to the limit state with high variance. Moreover, Zhang et al., [26] also propose the folded-normal distribution-based learning function called REIF and REIF2 (expected improvement function) to account for the modulating effect of the joint probability density function of input random variables on the scattering geometry of candidate samples. Follow this work, Shi et al.,[27] propose a novel learning function called Folded Normal based Expected Improvement Function (FNEIF) to precisely estimate the failure probability, which lies in an improvement function facilitating the prediction of surrogate model with folded normal variable, To flexibly enable enrichment of multiple training samples, parallel learning strategies such as k-means clustering-based approach proposed by Lelièvre et al. [28] and pseudo model-based method by Wen et al. [29] have been proposed. Considering many efficient learning strategies have been proposed by many researchers, it is worthy of the investigation of their computational efficiency for comparison. Despite the ground true for the definition of optimal learning strategy may not be available due various reasons, it is feasible to define the optimal learning strategy based on reasonable criteria.

In this paper, we explore the definition of optimal learning strategy. The goal is achieved by quantifying the variance of the estimated failure probability. Therefore, the optimal learning function is defined as the learning strategy that maximizes the variance reduction of failure probability after the selected training samples are enriched in current iterations. Concerning this, two optimal learning strategies with different level of time-complexities are proposed in this paper. Specifically, cases that considers and neglects the correlations among the candidate design samples are well elaborated. Moreover, we prove that the well-known *U* learning function can be reformulated to the general form of optimal learning function for the case neglecting the Kriging correlation. Moreover, we derive the mathematical expressions of Kriging correlation between two outputs, which facilitates the derivation of the optimal learning function. In addition, the theoretical optimal learning strategy for sequential multiple training samples enrichment is also mathematically explored through the Bayesian estimate with the corresponding lost functions. Simulation results show that the optimal learning strategy considering the Kriging correlation works better than that neglecting the Kriging correlation and other state-of-the art learning functions from the literatures in terms of the reduction of number of evaluations of performance function. However, the implementation of optimal learning strategy considering Kriging needs to investigate very large computational resource. Results showcase that the total number of evaluations to the performance function through enhanced parallel learning strategy can be even close to the conventional approaches through single training point enriching, while the total number of iterations via the former approach is significantly smaller than the later ones.

This article is organized as follows. Section 2 briefly introduces the elements for Kriging model and probabilistic classification-based Monte Carlo simulation. In Section 3, theorems regarding the optimal learning strategy without considering Kriging correlation are well elaborated. Section 4 derives the equations of Kriging correlation coefficient, which facilitates the optimal learning strategy considering the Kriging correlation. Moreover, section 5 proposes the theorem to derive the optimal parallel learning strategy with corresponding defined lost function. In section 6, two examples are investigated to demonstrate the proposed theorems. Section 7 discusses the algorithmic time-complexity of these propositions. This paper closes with conclusive remarks in section 8.

## 2. Adaptive Kriging-based Reliability Analysis

In Kriging surrogate model-based reliability analysis, the performance or limit state function, $g(\boldsymbol{x})$ is usually substituted by adaptively training a Kriging surrogate model $\hat{g}(\boldsymbol{x})$ [30], [31]. In this section, Kriging model with uncorrelated and correlated responses are briefly introduced [16], [22]. Generally, Kriging model $\hat{g}(\boldsymbol{x})$ can be represented as:

$$\hat{g}(\boldsymbol{x}) = F(\boldsymbol{x},\boldsymbol{\beta}) + Z(\boldsymbol{x}) = \boldsymbol{f}^T(\boldsymbol{x})\boldsymbol{\beta} + Z(\boldsymbol{x}), \tag{2}$$

where $F(\boldsymbol{x},\boldsymbol{\beta})$ is the regression base representing the Kriging trend, which can be a constant or a polynomial. $\boldsymbol{f}(\boldsymbol{x})$ is the Kriging basis and $\boldsymbol{\beta}$ is the regression coefficients. $\boldsymbol{f}^T(\boldsymbol{x})\boldsymbol{\beta}$ usually have ordinary ($\beta_0$), linear ($\beta_0$+$\sum_{n=1}^{N}\beta_n x_n$) or quadratic ($\beta_0$+$\sum_{n=1}^{N}\beta_n x_n$+$\sum_{n=1}^{N}\sum_{k=n}^{N}\beta_{nk}x_n x_k$) forms, whereas $n$ is the dimension of the random input vector, $\boldsymbol{x}$. Note that ordinary Kriging is used entirely in this paper. $Z(\boldsymbol{x})$ is the Kriging interpolation following a stationary normal Gaussian process with zero mean and a covariance matrix between two points, $\boldsymbol{x_i}$ and $\boldsymbol{x_j}$, as defined below.

$$\mathrm{COV}\left(Z(\boldsymbol{x}_i), Z(\boldsymbol{x}_j)\right) = \sigma^2 R(\boldsymbol{x_i}, \boldsymbol{x_j}; \boldsymbol{\theta}), \tag{3}$$

where $\sigma^2$ is the process variance or the generalized mean square error from the regression part and $R(\boldsymbol{x}_i, \boldsymbol{x}_j; \boldsymbol{\theta})$ is the correlation function or the kernel function representing the correlation function of the process with hyper-parameter $\boldsymbol{\theta}$. Multiple types of correlation functions are available in in Kriging model including linear, exponential, Gaussian, Matérn models and so on [31]. In this paper, the Gaussian kernel function is implemented:

$$R(\boldsymbol{x}_i, \boldsymbol{x}_j; \boldsymbol{\theta}) = \prod_{n=1}^{N} \exp\left(-\theta^n \left(x_i^n - x_j^n\right)^2\right), \tag{4}$$

where $N$ is the dimension of the random input vector, $\boldsymbol{x}_i$ or $\boldsymbol{x}_j$. The hyper-parameter $\boldsymbol{\theta}$ can be estimated via maximum likelihood estimation (MLE) or cross-validation [31]. It has been shown that the Kriging prediction is very sensitive to the value of $\boldsymbol{\theta}$ [22], [29], [32]. In this article, the well-known DACE toolbox is implemented [33], [34], [35], [36], where $\theta_i$ is searched in $(0,10)$. MLE is aimed to search for:

$$\boldsymbol{\theta}^* = \underset{\boldsymbol{\theta}\in\mathbb{R}}{\operatorname{argmin}}\left(\left|\boldsymbol{R}\left(\boldsymbol{x}_i, \boldsymbol{x}_j; \boldsymbol{\theta}\right)\right|^{\frac{1}{m}} \sigma^2\right), \tag{5}$$

where $m$ is the number of known training points or design-of-experiment (DoE) points. Thus, for a number of DoE points, $S_{DoE} = [\boldsymbol{x}_1, \boldsymbol{x}_2, \ldots, \boldsymbol{x}_m]$, and the corresponding responses from performance function $\mathbf{Y} = [G(\boldsymbol{x}_1), G(\boldsymbol{x}_2), \ldots, G(\boldsymbol{x}_m)]$, the traditional BLUP estimation of the Kriging prediction is as follows:

$$\mu_{\hat{g}}(\boldsymbol{x}) = \boldsymbol{f}^T(\boldsymbol{x})\boldsymbol{\beta} + \boldsymbol{r}(\boldsymbol{x})^T\boldsymbol{\gamma}, \tag{6}$$

Where

$$\begin{gathered}
\boldsymbol{\beta} = (\boldsymbol{F}^T\boldsymbol{R}^{-1}\boldsymbol{F})^{-1}\boldsymbol{F}^T\boldsymbol{R}^{-1}\boldsymbol{Y}, \\
\boldsymbol{\gamma} = \boldsymbol{R}^{-1}(\boldsymbol{Y} - \boldsymbol{F}\boldsymbol{\beta}), \\
\boldsymbol{r}(\boldsymbol{x}) = [R(\boldsymbol{x}_1, \boldsymbol{x}; \boldsymbol{\theta}), \ldots . R(\boldsymbol{x}_m, \boldsymbol{x}; \boldsymbol{\theta})]^T_{1\times m} \\
\boldsymbol{F} = [f(\boldsymbol{x}_1), f(\boldsymbol{x}_2), \ldots f(\boldsymbol{x}_m)]^T.
\end{gathered} \tag{7}$$

Then, the mean-square error (MSE) of $\hat{g}(\boldsymbol{x})$ can be calculated by:

$$\sigma^2_{\hat{g}}(\boldsymbol{x}) = \sigma^2(1 + \boldsymbol{u}^T(\boldsymbol{x})(\boldsymbol{F}^T\boldsymbol{R}^{-1}\boldsymbol{F})^{-1}\boldsymbol{u}(\boldsymbol{x}) - \boldsymbol{r}^T(\boldsymbol{x})\boldsymbol{R}^{-1}\boldsymbol{r}(\boldsymbol{x})), \tag{8}$$

where the Gaussian process variance and $\boldsymbol{u}(\boldsymbol{x})$ are:

$$\begin{gathered}
\sigma^2 = \frac{1}{m}(\boldsymbol{Y} - \boldsymbol{F}\boldsymbol{\beta})^T\boldsymbol{R}^{-1}(\boldsymbol{Y} - \boldsymbol{F}\boldsymbol{\beta}), \\
\boldsymbol{u}(\boldsymbol{x}) = \boldsymbol{F}^T\boldsymbol{R}^{-1}\boldsymbol{r}(\boldsymbol{x}) - \boldsymbol{f}(\boldsymbol{x}).
\end{gathered} \tag{9}$$

According to the Kriging model with uncorrelated responses, for all unknown points, $S_U = [\boldsymbol{x}^u_1, \boldsymbol{x}^u_2, \ldots, \boldsymbol{x}^u_{N_u}]$, the outputs $\boldsymbol{Y}_U = [G(\boldsymbol{x}^u_1), G(\boldsymbol{x}^u_2), \ldots, G(\boldsymbol{x}^u_{N_u})]$ from the Kriging are all mutually uncorrelated with the Kriging mean, $\mu_{\hat{g}}(\boldsymbol{x}^u_p)$, and the Kriging variance, $\sigma^2_{\hat{g}}(\boldsymbol{x}^u_p)$:

$$\boldsymbol{Y}_U(\boldsymbol{x}^u_p) \sim N\left(\mu_{\hat{g}}(\boldsymbol{x}^u_p), \sigma^2_{\hat{g}}(\boldsymbol{x}^u_p)\right), \quad \boldsymbol{x}^u_p \in S_U. \tag{10}$$

However, Kriging model with correlated responses can be represented as [22]:

$$\boldsymbol{Y}_U \sim N(\boldsymbol{\mu}_U, \boldsymbol{\Sigma}_U), \tag{11}$$

where the Kriging mean can be expressed in a matrix form:

$$\boldsymbol{\mu}_U = \boldsymbol{F}_U\boldsymbol{\beta} + \boldsymbol{r}^T_U\boldsymbol{R}^{-1}(\boldsymbol{Y} - \boldsymbol{F}\boldsymbol{\beta}), \tag{12}$$

and the corresponding covariance matrix can be read as:

$$\boldsymbol{\Sigma} = \sigma^2(\boldsymbol{R}_U + \boldsymbol{u}^T_U(\boldsymbol{F}^T\boldsymbol{R}^{-1}\boldsymbol{F})^{-1}\boldsymbol{u}_U - \boldsymbol{r}^T_U\boldsymbol{R}^{-1}\boldsymbol{r}_U), \tag{13}$$

Where

$$\begin{gathered}
\boldsymbol{R}_U = \left(\boldsymbol{R}\left(\boldsymbol{x}^u_p, \boldsymbol{x}^u_q; \boldsymbol{\theta}\right)\right)^T_{N_u\times N_u}, \boldsymbol{x}^u_p, \boldsymbol{x}^u_q \in S_U, \\
\boldsymbol{r}_U = \left(\boldsymbol{R}\left(\boldsymbol{x}_l, \boldsymbol{x}^u_p; \boldsymbol{\theta}\right)\right)^T_{N_u\times m}, \boldsymbol{x}_l \in S_{DoE},
\end{gathered}$$

$$\boldsymbol{u}_U = \boldsymbol{F}^T \boldsymbol{R}^{-1} \boldsymbol{r}_U - \boldsymbol{F}_U^T,$$
$$\boldsymbol{F}_U(\boldsymbol{x}) = \left[f(\boldsymbol{x}_1^u), f(\boldsymbol{x}_2^u), \dots f\left(\boldsymbol{x}_{N_u}^u\right)\right]^T. \tag{14}$$

Different from the Kriging model with uncorrelated responses, Kriging model with correlated responses follow the assumption in building the Kriging model that all the known and untried responses follow mutually correlated normal distribution. It is shown that Kriging model with correlated responses works better than that with uncorrelated responses in terms of estimating the confidence interval of failure probability [37], [38].

**2.2 Adaptive Kriging-based Reliability Methods**

Generally, probability of failure, denoted as $P_f$, can be expressed as,

$$P_f = \int_{\Omega_{g(x)\le 0}} f(\boldsymbol{x})\, dx = \int_{\Omega} I_g(\boldsymbol{x}) f(\boldsymbol{x}) d\boldsymbol{x}\,, \tag{15}$$

where $f(\boldsymbol{x})$ is the PDF of the random variables, $\boldsymbol{x}$, and $\Omega$ is the entire probabilistic domain. $\Omega_{g\le 0}$ is the integration domain where the performance function, $g(\boldsymbol{x}) \le 0$, and $I_g$ is the indicator function, with $I_g$=1, when $g(\boldsymbol{x}) \le 0$ and $I_g$=0 when $g(\boldsymbol{x}) > 0$. Because the true failure probability, $P_f$, is unavailable, an estimation based on probabilistic simulation techniques is preferred. Crude MCS with sufficiently large number of samples is often considered as a benchmark. In the Kriging model with MCS, there are two indicator functions: deterministic classification [16], [39] and the probabilistic classification [22], [40]. For deterministic classification, the crude MCS can be represented as:

$$\hat{P}_f^{dc} = \frac{1}{N_{MCS}} \sum_{i=1}^{N_{MCS}} I_g(\boldsymbol{x}_i)\,, \tag{16}$$

where $\hat{P}_f^{dc}$ denotes estimated failure probability with deterministic classification on the Kriging model. $N_{MCS}$ is the number of sampling realizations, $\boldsymbol{x}_i$, from the probabilistic distribution of random variables and $I_g(\cdot)$ is the corresponding indicator for deterministic classifications. The probabilistic classification-based failure probability can be estimated as:

$$\hat{P}_f^{pc} = \frac{1}{N_{MCS}} \sum_{i=1}^{N_{MCS}} I_{\hat{g}}(\boldsymbol{x}_i) = \frac{1}{N_{MCS}} \sum_{i=1}^{N_{MCS}} \Phi\left(\frac{-\mu_{\hat{g}}(\boldsymbol{x}_i)}{\sigma_{\hat{g}}(\boldsymbol{x}_i)}\right), \tag{17}$$

where $\hat{P}_f^{pc}$ is the estimated failure probability based on Kriging model and probabilistic classification for MCS. $\Phi(\cdot)$ is the cumulative distribution function (CDF) of the standard normal distribution and $\mu_{\hat{g}}(\boldsymbol{x}_i)$ and $\sigma_{\hat{g}}(\boldsymbol{x}_i)$ are the mean and standard deviation, respectively, of Kriging predictors. The probabilistic classification estimates the failure probability considering uncertainties associated with Kriging-based classification. In this paper, the probabilistic classification-based MCS is implemented since it is accurate [22] and also can provide the confidence interval for estimated failure probability according to [37]. Steps for adaptive Kriging-based reliability analysis can be briefly summarized in Algorithm. 1. For step 5, learning functions including EFF [17], U [16], LIF [24], H [23], REIF [26] and FNEIF [27] are investigated.

**Algorithm** 1. Adaptive Kriging-based Reliability Analysis

1. Generating initial candidate design samples $S$ with Latin Hypercube Sampling (LHS)
2. Randomly select initial training samples $\boldsymbol{x}_{tr}$ from $S$ and evaluate their responses $g(\boldsymbol{x}_{tr})$

3. Construct the Kriging model $\hat{g}(\boldsymbol{x})$ based on $\boldsymbol{x}_{tr}$ and $g(\boldsymbol{x}_{tr})$
4. Estimate the mean $\mu_{\hat{g}}(\boldsymbol{x})$, standard deviation $\sigma_{\hat{g}}(\boldsymbol{x})$ and $\hat{P}_f^{MCS}$ for $S$ with $\hat{g}(\boldsymbol{x})$
5. Search for the next best training points $\boldsymbol{x}_{tr}^*$ using learning function and update the training samples $\boldsymbol{x}_{tr}$
6. Check if the stopping criterion is satisfied or not:
   (a). Satisfied. Go to step 7.
   (b). Unsatisfied. Estimate the response $g(\boldsymbol{x}_{tr}^*)$ for $\boldsymbol{x}_{tr}^*$ and go back to Step 3.
7. Output $\hat{P}_f^{MCS}$

## 3. Optimal Learning strategies without considering Kriging correlation

Three learning strategies have shown substantial capability in strategically picking training samples for the surrogate construction. However, most of them are derived based on two factor that the selected point should be very close to the limit state as well as with large variance. Toward this goal, the theoretically optimal learning strategies are explored in this section. Essentially, the Kriging surrogate model is constructed based on the prior assumption that the outputs for design(training) and unknow samples follow multivariate Gaussian distribution with corresponding mean and covariance vector. Therefore, the Kriging correlation among training and unknown samples should be well considered. However, a good number of state-of-the-art researches for AK-based methodologies have also shown great computational efficiency even without considering the correlation. In this article, both the optimal learning strategies neglecting and considering Kriging correlation are analytically explored.

***Theorem 1***. Considering $\rho\left(\hat{y}(x_i), \hat{y}(x_j)\right) = 0$, the optimal active learning strategy can be expressed as follows:

$$\boldsymbol{x}_{tr}^* = \underset{\boldsymbol{x}_i \in S}{\arg\max}\ \Phi\left(\frac{-\mu_{\hat{g}}(\boldsymbol{x}_i)}{\sigma_{\hat{g}}(\boldsymbol{x}_i)}\right)\Phi\left(\frac{\mu_{\hat{g}}(\boldsymbol{x}_i)}{\sigma_{\hat{g}}(\boldsymbol{x}_i)}\right), i = 1,2, \dots N_{MCS} \tag{18}$$

Where $\rho\left(\hat{y}(x_i), \hat{y}(x_j)\right)$ denotes the Kriging correlation between two outputs, $\boldsymbol{x}_{tr}^*$ denotes the new training samples and $\Phi(\cdot)$ is the CDF of normal distribution.

**Proof**. Firstly, the fundamental issue one needs to solve is the definition of the optimal learning strategy. Considering the failure probability represented with stochastic estimator:

$$\tilde{P}_f^{mi} = \frac{1}{N_{MCS}}\mathbb{I}_{\Sigma}^{mi}, \qquad \boldsymbol{x}_i \in S \tag{19}$$

where $\tilde{P}_f^{mi}$ denotes the stochastic estimator of failure probability considering Kriging correlation, $N_{MCS}$ denotes the total number of candidate design samples and $\mathbb{I}_{\Sigma}^{mi}$ represents a random variable that is summed of multiple non-identically distributed and mutually independent(i.e., *mi* means mutually independent) Bernoulli random variables,

$$\mathbb{I}_{\Sigma}^{mi} = \sum_{i=1}^{N_{MCS}} I_{\hat{g}}^{mi}(\boldsymbol{x}_i), \boldsymbol{x}_i \in S \tag{20}$$

where $E[\cdot]$ is the expectation operator and $I_{\hat{g}}^{mi}(\boldsymbol{x}_i)$ is a Bernoulli random variable that can be expressed as follows:

$$I_{\hat{g}}^{mi}(\boldsymbol{x}_i) = \begin{cases} 1, & w.p\ \Phi\left(\frac{-\mu_{\hat{g}}(\boldsymbol{x}_i)}{\sigma_{\hat{g}}(\boldsymbol{x}_i)}\right) \\ 0, & w.p\ 1-\Phi\left(\frac{-\mu_{\hat{g}}(\boldsymbol{x}_i)}{\sigma_{\hat{g}}(\boldsymbol{x}_i)}\right) \end{cases}, \qquad \boldsymbol{x}_i \in S \tag{21}$$

where $\Phi(x)$ is the cumulative distribution function (CDF) of the univariate standard normal distribution. As Lindeberg's condition for the Central Limit Theorem for the sum of independent, non-identically distributed random variables [41] is satisfied for sufficiently large $N_{MCS}$:

$$\lim_{N_{MCS}\to\infty}\left(\max_{i=1,\dots,N_{MCS}}\frac{Var\left[I_{\hat{g}}^{mi}(\boldsymbol{x}_i)\right]}{Var\left[\mathbb{I}_{\Sigma}^{mi}\right]}\right)=0,\ \ \boldsymbol{x}_i\in\ S \tag{22}$$

it can be shown that $\mathbb{I}_{\Sigma}^{mi}$ in distribution converges to a normal distribution:

$$\mathbb{I}_{\Sigma}^{mi}\sim N\left(\mu_{\mathbb{I}_{\Sigma}^{mi}},\sigma_{\mathbb{I}_{\Sigma}^{mi}}^2\right),\qquad \boldsymbol{x}_i\in\ S. \tag{23}$$

Which indicates that:

$$\begin{aligned}\mathrm{Var}\left[\tilde{P}_f^{mi}\right]&=\mathrm{Var}\left[\frac{1}{N_{MCS}}\mathbb{I}_{\Sigma}^{mi}\right]=\frac{1}{N_{MCS}^2}\mathrm{Var}\left[\mathbb{I}_{\Sigma}^{mi}\right]\\&=\frac{1}{N_{MCS}^2}\mathrm{Var}\left[\sum_{i=1}^{N_{MCS}}I_{\hat{g}}^{mi}(\boldsymbol{x}_i)\right]=\frac{1}{N_{MCS}^2}\sum_{i=1}^{N_{MCS}}\mathrm{Var}\left[I_{\hat{g}}^{mi}(\boldsymbol{x}_i)\right]\\&=\frac{\sum_{i=1}^{N_{MCS}}\Phi\left(\frac{-\mu_{\hat{g}}(\boldsymbol{x}_i)}{\sigma_{\hat{g}}(\boldsymbol{x}_i)}\right)\left(1-\Phi\left(\frac{-\mu_{\hat{g}}(\boldsymbol{x}_i)}{\sigma_{\hat{g}}(\boldsymbol{x}_i)}\right)\right)}{N_{MCS}^2}\\&=\frac{\sum_{i=1}^{N_{MCS}}\Phi\left(\frac{-\mu_{\hat{g}}(\boldsymbol{x}_i)}{\sigma_{\hat{g}}(\boldsymbol{x}_i)}\right)\left(\Phi\left(\frac{\mu_{\hat{g}}(\boldsymbol{x}_i)}{\sigma_{\hat{g}}(\boldsymbol{x}_i)}\right)\right)}{N_{MCS}^2}\end{aligned} \tag{24}$$

As $\boldsymbol{x}_i$ is selected as the next training point, the following equation always holds true:

$$\lim_{\sigma_{\hat{g}}(x_i)\to 0}\left(\Phi\left(\frac{-\mu_{\hat{g}}(\boldsymbol{x}_i)}{\sigma_{\hat{g}}(\boldsymbol{x}_i)}\right)\left(1-\Phi\left(\frac{-\mu_{\hat{g}}(\boldsymbol{x}_i)}{\sigma_{\hat{g}}(\boldsymbol{x}_i)}\right)\right)\right)=0, \tag{25}$$

Let $\tilde{P}_f^{mi\prime}$ denote the stochastic estimator of failure probability after new training samples is enriched. Without considering the Kriging correlation, the optimal learning strategy can be represented as:

$$\begin{aligned}\boldsymbol{x}_{tr}^*&=\arg\max_{x_i\in\mathrm{S}}\ \left[\mathrm{Var}\left[\tilde{P}_f^{mi}\right]-\mathrm{Var}\left[\tilde{P}_f^{mi\prime}\right]\right],i=1,2,\dots N_{MCS}\\&=\arg\max_{x_i\in\mathrm{S}}\ \frac{1}{N_{MCS}^2}\left[\sum_{k=1}^{N_{MCS}}\Phi\left(\frac{-\mu_{\hat{g}}(\boldsymbol{x}_k)}{\sigma_{\hat{g}}(\boldsymbol{x}_k)}\right)\Phi\left(\frac{\mu_{\hat{g}}(\boldsymbol{x}_k)}{\sigma_{\hat{g}}(\boldsymbol{x}_k)}\right)\right.\\&\qquad\left.-\left(\sum_{k=1}^{N_{MCS}}\Phi\left(\frac{-\mu_{\hat{g}}(\boldsymbol{x}_k)}{\sigma_{\hat{g}}(\boldsymbol{x}_k)}\right)\Phi\left(\frac{\mu_{\hat{g}}(\boldsymbol{x}_k)}{\sigma_{\hat{g}}(\boldsymbol{x}_k)}\right)-\Phi\left(\frac{-\mu_{\hat{g}}(\boldsymbol{x}_i)}{\sigma_{\hat{g}}(\boldsymbol{x}_i)}\right)\Phi\left(\frac{\mu_{\hat{g}}(\boldsymbol{x}_i)}{\sigma_{\hat{g}}(\boldsymbol{x}_i)}\right)\right)\right],i=1,2,\dots N_{MCS}\\&=\arg\max_{x_i\in\mathrm{S}}\ \Phi\left(\frac{-\mu_{\hat{g}}(\boldsymbol{x}_i)}{\sigma_{\hat{g}}(\boldsymbol{x}_i)}\right)\Phi\left(\frac{\mu_{\hat{g}}(\boldsymbol{x}_i)}{\sigma_{\hat{g}}(\boldsymbol{x}_i)}\right),i=1,2,\dots N_{MCS}\end{aligned} \tag{26}$$

This shows that the learning strategy without considering Kriging correlation tends to select the one that make the most contributions to the value of variance of $\tilde{P}_f^{mi}$. In this case, the value of $v(\boldsymbol{x}_i)$ is actually equal to $v(\boldsymbol{x}_i) = \Phi\left(\frac{-\mu_{\hat{g}}(\boldsymbol{x}_i)}{\sigma_{\hat{g}}(\boldsymbol{x}_i)}\right)\Phi\left(\frac{\mu_{\hat{g}}(\boldsymbol{x}_i)}{\sigma_{\hat{g}}(\boldsymbol{x}_i)}\right)$.

***Lemma 1***. Considering theorem 1, the *U* learning function is the optimal learning strategy. This infers that $\mathcal{T}(\boldsymbol{x}_{tr}^*) = 0$, where $\mathcal{T}(\boldsymbol{x}_{tr}^*)$ can be expressed as follows:

$$\mathcal{T}(\boldsymbol{x}_{tr}^*) = \Phi\left(\frac{-\mu_{\hat{g}}(\boldsymbol{x}_{tr}^*)}{\sigma_{\hat{g}}(\boldsymbol{x}_{tr}^*)}\right)\Phi\left(\frac{\mu_{\hat{g}}(\boldsymbol{x}_{tr}^*)}{\sigma_{\hat{g}}(\boldsymbol{x}_{tr}^*)}\right) - \int\left(U(\boldsymbol{x}_{tr}^*)\right) \tag{27}$$

Here $\int(\cdot)$ denotes a monotonical function.

**Proof.** Let $\mathcal{T}(\boldsymbol{x}_{tr}^*) = 0$, which indicates:

$$\int\left(U(\boldsymbol{x}_{tr}^*)\right) = \Phi\left(\frac{-\mu_{\hat{g}}(\boldsymbol{x}_{tr}^*)}{\sigma_{\hat{g}}(\boldsymbol{x}_{tr}^*)}\right)\Phi\left(\frac{\mu_{\hat{g}}(\boldsymbol{x}_{tr}^*)}{\sigma_{\hat{g}}(\boldsymbol{x}_{tr}^*)}\right) \tag{28}$$

with $U(\boldsymbol{x}_{tr}^*) = \frac{|\mu_{\hat{g}}(\boldsymbol{x}_{tr}^*)|}{\sigma_{\hat{g}}(\boldsymbol{x}_{tr}^*)}$ and

$$\Phi\left(\frac{-\mu_{\hat{g}}(\boldsymbol{x}_{tr}^*)}{\sigma_{\hat{g}}(\boldsymbol{x}_{tr}^*)}\right)\Phi\left(\frac{\mu_{\hat{g}}(\boldsymbol{x}_{tr}^*)}{\sigma_{\hat{g}}(\boldsymbol{x}_{tr}^*)}\right) = \Phi(-U)\Phi(U) = \left(1-\Phi(U)\right)\Phi(U) = \Phi(U) - \Phi^2(U) \tag{29}$$

One can get:

$$\int(U) = \Phi(U) - \Phi^2(U), U \geq 0 \tag{30}$$

Moreover,

$$\frac{\partial \int(U)}{\partial \Phi(U)} = 1 - \frac{\Phi(U)}{2} \leq 0, \Phi(U) \in [0.5,1] \tag{31}$$

This implies that $\int(v)$ is a monotonically function, and the '*U*' learning function is equivalent to the optimal learning strategy without considering Kriging correlation. Without losing generality, any learning functions that satisfy Eq. (27) are among one of the mathematical forms of Eq. (18). In the following section, the optimal learning strategy that considering the Kriging correlation is analytically derived.

## 4. Optimal Learning strategies considering Kriging correlation

The above represented optimal learning strategies rely on the assumption that the Kriging outputs are mutually independent. As the Kriging correlation is considered, the optimal learning strategies should change accordingly. Reconsider Eq. (15) ~ (17), the failure probability can be represented as a stochastic indicator:

$$\tilde{P}_f^{mc} = \frac{1}{N_{MCS}} \mathbb{I}_{\Sigma}^{mc}, \qquad \boldsymbol{x}_i \in S \tag{32}$$

where $\tilde{P}_f^{mc}$ denotes the stochastic estimator of failure probability considering Kriging correlation and $\mathbb{I}_{\Sigma}^{mc}$ represents a random variable that is summed of multiple non-identically distributed and mutually correlated (i.e., *mc* means mutually correlated) Bernoulli random variables,

$$\mathbb{I}_{\Sigma}^{mc} = \sum_{i=1}^{N_{MCS}} I_{\hat{g}}^{mc}(\boldsymbol{x}_i)\,, \boldsymbol{x}_i \in S \tag{33}$$

where $I_{\hat{g}}^{mc}(\boldsymbol{x}_i)$ is a correlated Bernoulli random variable with $I_{\hat{g}}^{mc}(\boldsymbol{x}_i) = 1$ denoting failure, and the mean and covariance matrix of correlated Bernoulli distribution $\boldsymbol{B} = \left[I_{\hat{g}}^{mc}(\boldsymbol{x}_1), I_{\hat{g}}^{mc}(\boldsymbol{x}_2), \dots I_{\hat{g}}^{mc}(\boldsymbol{x}_{N_{MCS}})\right]$ be represented as:

$$\boldsymbol{\mu}_b = \begin{bmatrix} \mu_b(\boldsymbol{x}_1) & 0 & \cdots & 0 \\ 0 & \mu_b(\boldsymbol{x}_2) & & 0 \\ \vdots & & \ddots & \vdots \\ 0 & 0 & \cdots & \mu_b(\boldsymbol{x}_{N_{MCS}}) \end{bmatrix}, \boldsymbol{x}_i \in S \tag{34}$$

and

$$\boldsymbol{\Sigma}_b = \begin{bmatrix} {\sigma_b}^2(\boldsymbol{x}_1) & \rho_{1,2}\sigma_b(\boldsymbol{x}_1)\sigma_b(\boldsymbol{x}_2) & \cdots & \rho_{1,N_{MCS}}\sigma_b(\boldsymbol{x}_1)\sigma_b(\boldsymbol{x}_{N_{MCS}}) \\ \rho_{2,1}\sigma_b(\boldsymbol{x}_2)\sigma_b(\boldsymbol{x}_1) & {\sigma_b}^2(\boldsymbol{x}_2) & & \rho_{2,N_{MCS}}\sigma_b(\boldsymbol{x}_2)\sigma_b(\boldsymbol{x}_{N_{MCS}}) \\ \vdots & & \ddots & \vdots \\ \rho_{N_{MCS},1}\sigma_b(\boldsymbol{x}_{N_{MCS}})\sigma_b(\boldsymbol{x}_1) & \rho_{N_{MCS},2}\sigma_b(\boldsymbol{x}_{N_{MCS}})\sigma_b(\boldsymbol{x}_2) & \cdots & {\sigma_b}^2(\boldsymbol{x}_{N_{MCS}}) \end{bmatrix}, \boldsymbol{x}_i \in S \tag{35}$$

where ${\sigma_b}^2(\boldsymbol{x}_i)$ is the Bernoulli variance and $\rho_{i,j}$ is the corresponding correlation of two Bernoulli random variables $\boldsymbol{x}_i$ and $\boldsymbol{x}_j$. Moreover, ${\sigma_b}^2(\boldsymbol{x}_i)$ can be calculated as:

$$\begin{aligned} {\sigma_b}^2(\boldsymbol{x}_i) &= \Phi\left(\frac{-\mu_{\hat{g}}(\boldsymbol{x}_i)}{\sigma_{\hat{g}}(\boldsymbol{x}_i)}\right)\cdot\left(1 - \Phi\left(\frac{-\mu_{\hat{g}}(\boldsymbol{x}_i)}{\sigma_{\hat{g}}(\boldsymbol{x}_i)}\right)\right) \\ &= \Phi\left(\frac{-\mu_{\hat{g}}(\boldsymbol{x}_i)}{\sigma_{\hat{g}}(\boldsymbol{x}_i)}\right)\cdot\Phi\left(\frac{\mu_{\hat{g}}(\boldsymbol{x}_i)}{\sigma_{\hat{g}}(\boldsymbol{x}_i)}\right), \boldsymbol{x}_i \in S. \end{aligned} \tag{36}$$

Therefore, the probability that $\boldsymbol{B} = [1,1,\dots,1]_{N_{MCS}\times 1}$ can be represented as:

$$\begin{aligned} \Phi_{b=1}(\boldsymbol{x}_1, \boldsymbol{x}_2, \dots \boldsymbol{x}_{N_{MCS}}) &= \int_{-\infty}^{0} \dots \int_{-\infty}^{0} \varphi\left(\left[x_1, \dots, x_{N_{MCS}}\right]_{N_{MCS}\times 1}; \boldsymbol{\mu}_{\hat{g}}, \boldsymbol{\Sigma}_{\hat{g}}\right) dx_1 \dots dx_{N_{MCS}} \\ &= \Phi\left([0,\dots,0]_{N_{MCS}\times 1}; \boldsymbol{\mu}_{\hat{g}}, \boldsymbol{\Sigma}_{\hat{g}}\right), i = 1,2,3, \dots N_{MCS} \end{aligned} \tag{37}$$

where $\Phi_{b=1}$ denotes the probability of $\boldsymbol{B} = [1,1,\dots,1]_{N_{MCS}\times 1}$, $\Phi(\cdot)$ and $\varphi(\cdot)$ are the CDF and PDF of the multivariate normal distribution, respectively, with mean $\boldsymbol{\mu}_{\hat{g}}$ and covariance matrix $\boldsymbol{\Sigma}_{\hat{g}}$ of Kriging output. According to the Central Limit Theorem for correlated random variables [42], the variance of the stochastic estimator can be estimated as:

$$\mathrm{Var}\left[\tilde{P}_f^{mc}\right] = \mathrm{Var}\left[\frac{1}{N_{MCS}}\mathbb{I}_{\Sigma}^{mc}\right] = \frac{1}{N_{MCS}^2}\mathrm{Var}\left[\mathbb{I}_{\Sigma}^{mi}\right] = \frac{1}{N_{MCS}^2}\sum_{j=1}^{N_{MCS}}\sum_{k=1}^{N_{MCS}} \boldsymbol{\Sigma}_{b\,j,k} \tag{38}$$

where $\Sigma_{b\,j,k}$ denotes the j[th] row and k[th] column of the elements in covariance matrix $\boldsymbol{\Sigma}_b$. The covariance between two correlated Bernoulli random variable $\boldsymbol{x}_i$ and $\boldsymbol{x}_j$ can be represented as:

$$COV\left(I_{\hat{g}}^{mc}(\boldsymbol{x}_i), I_{\hat{g}}^{mc}(\boldsymbol{x}_i)\right) = \rho_{i,j}^{b} \cdot \sigma_b(\boldsymbol{x}_i) \cdot \sigma_b(\boldsymbol{x}_j), \tag{39}$$

where $\rho_{i,j}^{b} \in [-1,1]$ is an unknown correlation coefficient of two correlated Bernoulli distributions, $I_{\hat{g}}(\boldsymbol{x}_i)$ and $I_{\hat{g}}(\boldsymbol{x}_j)$. $\sigma_b(\boldsymbol{x}_i)$ and $\sigma_b(\boldsymbol{x}_j)$ are the standard deviations ($\sigma_b(\boldsymbol{x}_i) = \sqrt{\mu_b(\boldsymbol{x}_i)(1-\mu_b(\boldsymbol{x}_i))}$).

***Theorem 2***. The correlation between the two correlated Bernoulli random variables $I_{\hat{g}}^{mc}(\boldsymbol{x}_i)$ and $I_{\hat{g}}^{mc}(\boldsymbol{x}_i)$ can be calculated as follows:

$$\rho_{i,j}^{b} = \frac{\Phi\left([0,0];[\mu_i;\mu_j],\begin{bmatrix}\Sigma_{i,i} & \Sigma_{j,i}\\ \Sigma_{i,j} & \Sigma_{j,j}\end{bmatrix}\right) - \Phi\left(\frac{-\mu_{\hat{g}}(\boldsymbol{x}_i)}{\sigma_{\hat{g}}(\boldsymbol{x}_i)}\right)\Phi\left(\frac{-\mu_{\hat{g}}(\boldsymbol{x}_j)}{\sigma_{\hat{g}}(\boldsymbol{x}_j)}\right)}{\sqrt{\Phi\left(\frac{-\mu_{\hat{g}}(\boldsymbol{x}_i)}{\sigma_{\hat{g}}(\boldsymbol{x}_i)}\right)\cdot\Phi\left(\frac{\mu_{\hat{g}}(\boldsymbol{x}_i)}{\sigma_{\hat{g}}(\boldsymbol{x}_i)}\right)\,\Phi\left(\frac{-\mu_{\hat{g}}(\boldsymbol{x}_j)}{\sigma_{\hat{g}}(\boldsymbol{x}_j)}\right)\cdot\Phi\left(\frac{\mu_{\hat{g}}(\boldsymbol{x}_j)}{\sigma_{\hat{g}}(\boldsymbol{x}_j)}\right)}} \tag{40}$$

where $\mu_i$ and $\mu_j$ denotes the i[th] and j[th] elements of the mean matrix $\boldsymbol{\mu}_{\hat{g}}$ and $\Sigma_{i,j}$ denotes the i[th] row and j[th] column of the covariance matrix $\boldsymbol{\Sigma}_{\hat{g}}$, which can be reprenseted as:

$$\boldsymbol{\mu}_{\hat{g}} = \begin{bmatrix}\mu_{\hat{g}}(\boldsymbol{x}_1) & 0 & \cdots & 0\\ 0 & \mu_{\hat{g}}(\boldsymbol{x}_2) & & 0\\ \vdots & & \ddots & \vdots\\ 0 & 0 & \cdots & \mu_{\hat{g}}(\boldsymbol{x}_{N_{MCS}})\end{bmatrix}, \boldsymbol{x}_i \in S \tag{41}$$

and

$$\boldsymbol{\Sigma}_{\hat{g}} = \begin{bmatrix}\sigma_{\hat{g}}^2(\boldsymbol{x}_1) & \rho_{1,2}\sigma_{\hat{g}}(\boldsymbol{x}_1)\sigma_{\hat{g}}(\boldsymbol{x}_2) & \cdots & \rho_{1,N_{MCS}}\sigma_{\hat{g}}(\boldsymbol{x}_1)\sigma_{\hat{g}}(\boldsymbol{x}_{N_{MCS}})\\ \rho_{2,1}\sigma_{\hat{g}}(\boldsymbol{x}_2)\sigma_{\hat{g}}(\boldsymbol{x}_1) & \sigma_{\hat{g}}^2(\boldsymbol{x}_2) & & \rho_{2,N_{MCS}}\sigma_{\hat{g}}(\boldsymbol{x}_2)\sigma_{\hat{g}}(\boldsymbol{x}_{N_{MCS}})\\ \vdots & & \ddots & \vdots\\ \rho_{N_{MCS},1}\sigma_{\hat{g}}(\boldsymbol{x}_{N_{MCS}})\sigma_{\hat{g}}(\boldsymbol{x}_1) & \rho_{N_{MCS},2}\sigma_{\hat{g}}(\boldsymbol{x}_{N_{MCS}})\sigma_{\hat{g}}(\boldsymbol{x}_2) & \cdots & \sigma_{\hat{g}}^2(\boldsymbol{x}_{N_{MCS}})\end{bmatrix}, \tag{42}$$

**Proof**: The correlation function can be estimated as follows

$$\rho_{i,j}^{b} = \frac{CoV\left(I_{\hat{g}}^{mc}(\boldsymbol{x}_i), I_{\hat{g}}^{mc}(\boldsymbol{x}_i)\right)}{\sigma_b(\boldsymbol{x}_i)\cdot\sigma_b(\boldsymbol{x}_j)} \tag{43}$$

Given that,

$$\begin{aligned} CoV\left(I_{\hat{g}}^{mc}(\boldsymbol{x}_i), I_{\hat{g}}^{mc}(\boldsymbol{x}_i)\right) &= E\left[\left(I_{\hat{g}}^{mc}(\boldsymbol{x}_i)-\mu_b(\boldsymbol{x}_i)\right)\left(I_{\hat{g}}^{mc}(\boldsymbol{x}_j)-\mu_b(\boldsymbol{x}_j)\right)\right]\\ &= E\left(I_{\hat{g}}^{mc}(\boldsymbol{x}_i)I_{\hat{g}}^{mc}(\boldsymbol{x}_j)\right) - E\left(I_{\hat{g}}^{mc}(\boldsymbol{x}_i)\right)E\left(I_{\hat{g}}^{mc}(\boldsymbol{x}_j)\right)\\ &= \int_{-\infty}^{0}\int_{-\infty}^{0}\varphi\left([x,y];[\mu_i;\mu_j],\begin{bmatrix}\Sigma_{i,i} & \Sigma_{j,i}\\ \Sigma_{i,j} & \Sigma_{j,j}\end{bmatrix}\right)dxdy - \Phi\left(\frac{-\mu_{\hat{g}}(\boldsymbol{x}_i)}{\sigma_{\hat{g}}(\boldsymbol{x}_i)}\right)\Phi\left(\frac{-\mu_{\hat{g}}(\boldsymbol{x}_j)}{\sigma_{\hat{g}}(\boldsymbol{x}_j)}\right)\\ &= \Phi\left([0,0];[\mu_i;\mu_j],\begin{bmatrix}\Sigma_{i,i} & \Sigma_{j,i}\\ \Sigma_{i,j} & \Sigma_{j,j}\end{bmatrix}\right) - \Phi\left(\frac{-\mu_{\hat{g}}(\boldsymbol{x}_i)}{\sigma_{\hat{g}}(\boldsymbol{x}_i)}\right)\Phi\left(\frac{-\mu_{\hat{g}}(\boldsymbol{x}_j)}{\sigma_{\hat{g}}(\boldsymbol{x}_j)}\right) \end{aligned} \tag{44}$$

And

$$\sigma_b(\boldsymbol{x}_i)\cdot\sigma_b(\boldsymbol{x}_j)=\sqrt{\sigma_b{}^2(\boldsymbol{x}_i)\sigma_b{}^2(\boldsymbol{x}_j)}$$
$$=\sqrt{\Phi\left(\frac{-\mu_{\hat{g}}(\boldsymbol{x}_i)}{\sigma_{\hat{g}}(\boldsymbol{x}_i)}\right)\cdot\Phi\left(\frac{\mu_{\hat{g}}(\boldsymbol{x}_i)}{\sigma_{\hat{g}}(\boldsymbol{x}_i)}\right)\Phi\left(\frac{-\mu_{\hat{g}}(\boldsymbol{x}_j)}{\sigma_{\hat{g}}(\boldsymbol{x}_j)}\right)\cdot\Phi\left(\frac{\mu_{\hat{g}}(\boldsymbol{x}_j)}{\sigma_{\hat{g}}(\boldsymbol{x}_j)}\right)} \tag{45}$$

Thus, the correlation function can be estimated as:

$$\rho_{i,j}^b=\frac{CoV\left(I_{\hat{g}}^{mc}(\boldsymbol{x}_i),I_{\hat{g}}^{mc}(\boldsymbol{x}_j)\right)}{\sigma_b(\boldsymbol{x}_i)\cdot\sigma_b(\boldsymbol{x}_j)}=\frac{\Phi\left([0,0];[\mu_i;\mu_j],\begin{bmatrix}\Sigma_{i,i} & \Sigma_{j,i}\\ \Sigma_{i,j} & \Sigma_{j,j}\end{bmatrix}\right)-\Phi\left(\frac{-\mu_{\hat{g}}(\boldsymbol{x}_i)}{\sigma_{\hat{g}}(\boldsymbol{x}_i)}\right)\Phi\left(\frac{-\mu_{\hat{g}}(\boldsymbol{x}_j)}{\sigma_{\hat{g}}(\boldsymbol{x}_j)}\right)}{\sqrt{\Phi\left(\frac{-\mu_{\hat{g}}(\boldsymbol{x}_i)}{\sigma_{\hat{g}}(\boldsymbol{x}_i)}\right)\cdot\Phi\left(\frac{\mu_{\hat{g}}(\boldsymbol{x}_i)}{\sigma_{\hat{g}}(\boldsymbol{x}_i)}\right)\Phi\left(\frac{-\mu_{\hat{g}}(\boldsymbol{x}_j)}{\sigma_{\hat{g}}(\boldsymbol{x}_j)}\right)\cdot\Phi\left(\frac{\mu_{\hat{g}}(\boldsymbol{x}_j)}{\sigma_{\hat{g}}(\boldsymbol{x}_j)}\right)}} \tag{46}$$

***Remark***. Specifically, as $\rho_{i,j}=0$, it means that the failure probabilities of two points, $\boldsymbol{x}_i$ and $\boldsymbol{x}_j$, are uncorrelated, which can further infer that $\rho_{i,j}^b=0$.

$$\rho_{i,j}^b=\frac{\Phi\left([0,0];[\mu_i;\mu_j],\begin{bmatrix}\Sigma_{i,i} & 0\\ 0 & \Sigma_{j,j}\end{bmatrix}\right)-\Phi\left(\frac{-\mu_{\hat{g}}(\boldsymbol{x}_i)}{\sigma_{\hat{g}}(\boldsymbol{x}_i)}\right)\Phi\left(\frac{-\mu_{\hat{g}}(\boldsymbol{x}_j)}{\sigma_{\hat{g}}(\boldsymbol{x}_j)}\right)}{\sqrt{\Phi\left(\frac{-\mu_{\hat{g}}(\boldsymbol{x}_i)}{\sigma_{\hat{g}}(\boldsymbol{x}_i)}\right)\cdot\Phi\left(\frac{\mu_{\hat{g}}(\boldsymbol{x}_i)}{\sigma_{\hat{g}}(\boldsymbol{x}_i)}\right)\Phi\left(\frac{-\mu_{\hat{g}}(\boldsymbol{x}_j)}{\sigma_{\hat{g}}(\boldsymbol{x}_j)}\right)\cdot\Phi\left(\frac{\mu_{\hat{g}}(\boldsymbol{x}_j)}{\sigma_{\hat{g}}(\boldsymbol{x}_j)}\right)}}$$
$$=\frac{\Phi\left(\frac{-\mu_{\hat{g}}(\boldsymbol{x}_i)}{\sigma_{\hat{g}}(\boldsymbol{x}_i)}\right)\Phi\left(\frac{-\mu_{\hat{g}}(\boldsymbol{x}_j)}{\sigma_{\hat{g}}(\boldsymbol{x}_j)}\right)-\Phi\left(\frac{-\mu_{\hat{g}}(\boldsymbol{x}_i)}{\sigma_{\hat{g}}(\boldsymbol{x}_i)}\right)\Phi\left(\frac{-\mu_{\hat{g}}(\boldsymbol{x}_j)}{\sigma_{\hat{g}}(\boldsymbol{x}_j)}\right)}{\sqrt{\Phi\left(\frac{-\mu_{\hat{g}}(\boldsymbol{x}_i)}{\sigma_{\hat{g}}(\boldsymbol{x}_i)}\right)\cdot\Phi\left(\frac{\mu_{\hat{g}}(\boldsymbol{x}_i)}{\sigma_{\hat{g}}(\boldsymbol{x}_i)}\right)\Phi\left(\frac{-\mu_{\hat{g}}(\boldsymbol{x}_j)}{\sigma_{\hat{g}}(\boldsymbol{x}_j)}\right)\cdot\Phi\left(\frac{\mu_{\hat{g}}(\boldsymbol{x}_j)}{\sigma_{\hat{g}}(\boldsymbol{x}_j)}\right)}}=0 \tag{47}$$

When $\rho_{i,j}=1$, it means that the two responses are positively correlated. Concerning $\Phi\left([0,0];[\mu_i;\mu_j],\begin{bmatrix}\Sigma_{i,i} & -1\\ -1 & \Sigma_{j,j}\end{bmatrix}\right)=\Phi\left(\frac{-\mu_{\hat{g}}(x_i)}{\sigma_{\hat{g}}(x_i)}\right)$ and $\Phi\left(\frac{-\mu_{\hat{g}}(x_i)}{\sigma_{\hat{g}}(x_i)}\right)=\Phi\left(\frac{-\mu_{\hat{g}}(x_j)}{\sigma_{\hat{g}}(x_j)}\right)$,

$$\rho_{i,j}^b=\frac{\Phi\left([0,0];[\mu_i;\mu_j],\begin{bmatrix}\Sigma_{i,i} & 1\\ 1 & \Sigma_{j,j}\end{bmatrix}\right)-\Phi\left(\frac{-\mu_{\hat{g}}(\boldsymbol{x}_i)}{\sigma_{\hat{g}}(\boldsymbol{x}_i)}\right)\Phi\left(\frac{-\mu_{\hat{g}}(\boldsymbol{x}_j)}{\sigma_{\hat{g}}(\boldsymbol{x}_j)}\right)}{\sqrt{\Phi\left(\frac{-\mu_{\hat{g}}(\boldsymbol{x}_i)}{\sigma_{\hat{g}}(\boldsymbol{x}_i)}\right)\cdot\Phi\left(\frac{\mu_{\hat{g}}(\boldsymbol{x}_i)}{\sigma_{\hat{g}}(\boldsymbol{x}_i)}\right)\Phi\left(\frac{-\mu_{\hat{g}}(\boldsymbol{x}_j)}{\sigma_{\hat{g}}(\boldsymbol{x}_j)}\right)\cdot\Phi\left(\frac{\mu_{\hat{g}}(\boldsymbol{x}_j)}{\sigma_{\hat{g}}(\boldsymbol{x}_j)}\right)}}$$
$$=\frac{\Phi\left(\frac{-\mu_{\hat{g}}(\boldsymbol{x}_i)}{\sigma_{\hat{g}}(\boldsymbol{x}_i)}\right)-\Phi\left(\frac{-\mu_{\hat{g}}(\boldsymbol{x}_i)}{\sigma_{\hat{g}}(\boldsymbol{x}_i)}\right)\Phi\left(\frac{-\mu_{\hat{g}}(\boldsymbol{x}_i)}{\sigma_{\hat{g}}(\boldsymbol{x}_i)}\right)}{\Phi\left(\frac{-\mu_{\hat{g}}(\boldsymbol{x}_i)}{\sigma_{\hat{g}}(\boldsymbol{x}_i)}\right)\cdot\Phi\left(\frac{\mu_{\hat{g}}(\boldsymbol{x}_i)}{\sigma_{\hat{g}}(\boldsymbol{x}_i)}\right)}=1 \tag{48}$$

This also applies to the case that $\rho_{i,j}=-1$ with $\Phi\left([0,0];[\mu_i;\mu_j],\begin{bmatrix}\Sigma_{i,i} & -1\\ -1 & \Sigma_{j,j}\end{bmatrix}\right)=0$ and $\Phi\left(\frac{-\mu_{\hat{g}}(x_i)}{\sigma_{\hat{g}}(x_i)}\right)=\Phi\left(\frac{\mu_{\hat{g}}(x_j)}{\sigma_{\hat{g}}(x_j)}\right)$:

$$\rho_{i,j}^{b} = \frac{\Phi\left([0,0];[\mu_i;\mu_j],\begin{bmatrix}\Sigma_{i,i} & -1\\ -1 & \Sigma_{j,j}\end{bmatrix}\right) - \Phi\left(\frac{-\mu_{\hat{g}}(\boldsymbol{x}_i)}{\sigma_{\hat{g}}(\boldsymbol{x}_i)}\right)\Phi\left(\frac{-\mu_{\hat{g}}(\boldsymbol{x}_j)}{\sigma_{\hat{g}}(\boldsymbol{x}_j)}\right)}{\sqrt{\Phi\left(\frac{-\mu_{\hat{g}}(\boldsymbol{x}_i)}{\sigma_{\hat{g}}(\boldsymbol{x}_i)}\right)\cdot\Phi\left(\frac{\mu_{\hat{g}}(\boldsymbol{x}_i)}{\sigma_{\hat{g}}(\boldsymbol{x}_i)}\right)\Phi\left(\frac{-\mu_{\hat{g}}(\boldsymbol{x}_j)}{\sigma_{\hat{g}}(\boldsymbol{x}_j)}\right)\cdot\Phi\left(\frac{\mu_{\hat{g}}(\boldsymbol{x}_j)}{\sigma_{\hat{g}}(\boldsymbol{x}_j)}\right)}}$$

$$= \frac{-\Phi\left(\frac{-\mu_{\hat{g}}(\boldsymbol{x}_i)}{\sigma_{\hat{g}}(\boldsymbol{x}_i)}\right)\Phi\left(\frac{\mu_{\hat{g}}(\boldsymbol{x}_i)}{\sigma_{\hat{g}}(\boldsymbol{x}_i)}\right)}{\Phi\left(\frac{-\mu_{\hat{g}}(\boldsymbol{x}_i)}{\sigma_{\hat{g}}(\boldsymbol{x}_i)}\right)\cdot\Phi\left(\frac{\mu_{\hat{g}}(\boldsymbol{x}_i)}{\sigma_{\hat{g}}(\boldsymbol{x}_i)}\right)} = -1 \tag{49}$$

Therefore, it can further infer that $\rho_{i,j}^{b} \propto \rho_{i,j}$. Moreover, aforementioned contents facilitate the derivation of optimal learning strategy considering the Kriging correlation. Computational details regarding Eq. (40) are elaborated in the Appendix A2.

***Theorem 3***. Considering $\rho\left(\hat{y}(x_i),\hat{y}(x_j)\right) \neq 0$, the optimal active learning strategy can be expressed by the equation below:

$$\boldsymbol{x}_{tr}^{*} = \underset{x_i \in \mathrm{S}}{\arg\max}\left\{2\sum_{k=1}^{N_{MCS}}\left(\rho_{i,k}^{b}\sigma_b(\boldsymbol{x}_i)\sigma_b(\boldsymbol{x}_k)\right) - {\sigma_b}^2(\boldsymbol{x}_i)\right\}, i = 1,2,\ldots N_{MCS} \tag{50}$$

Where $\sigma_b(\boldsymbol{x}) = \Phi\left(\frac{-\mu_{\hat{g}}(x)}{\sigma_{\hat{g}}(x)}\right)\cdot\Phi\left(\frac{\mu_{\hat{g}}(x)}{\sigma_{\hat{g}}(x)}\right)$ and $\rho_{i,k}^{b}$ can be calculated using Eq. (40).

**Proof**. Given that,

$$\lim_{\sigma_{\hat{g}}(x_i)\to 0}\Sigma_{b\,i,k} = \lim_{\sigma_{\hat{g}}(x_i)\to 0}\rho_{i,k}\sqrt{\Phi\left(\frac{-\mu_{\hat{g}}(\boldsymbol{x}_i)}{\sigma_{\hat{g}}(\boldsymbol{x}_i)}\right)\cdot\left(1-\Phi\left(\frac{-\mu_{\hat{g}}(\boldsymbol{x}_i)}{\sigma_{\hat{g}}(\boldsymbol{x}_i)}\right)\right)\Phi\left(\frac{-\mu_{\hat{g}}(\boldsymbol{x}_k)}{\sigma_{\hat{g}}(\boldsymbol{x}_k)}\right)\cdot\left(1-\Phi\left(\frac{-\mu_{\hat{g}}(\boldsymbol{x}_k)}{\sigma_{\hat{g}}(\boldsymbol{x}_k)}\right)\right)} \tag{51}$$

$$= 0, \mathrm{k} = 1,2,\ldots N_{MCS}$$

Following the same principle in section 3, the optima learning strategy can be represented as:

$$\boldsymbol{x}_{tr}^{*} = \underset{x_i \in \mathrm{S}}{\arg\max}\left[\mathrm{Var}\left[\tilde{P}_f^{mc}\right] - \mathrm{Var}\left[\tilde{P}_f^{mc'}\right]\right]$$

$$= \frac{1}{N_{MCS}^2}\underset{x_i \in \mathrm{S}}{\arg\max}\left[\sum_{j=1}^{N_{MCS}}\sum_{k=1}^{N_{MCS}}\boldsymbol{\Sigma}_{b\,j,k} - \left(\sum_{j=1}^{N_{MCS}}\sum_{k=1}^{N_{MCS}}\boldsymbol{\Sigma}_{b\,j,k} - \left(\sum_{i=1}^{N_{MCS}}\left(\boldsymbol{\Sigma}_{pb\,i,k}\right) + \sum_{j=1}^{N_{MCS}}\left(\boldsymbol{\Sigma}_{pb\,k,i}\right) - {\sigma_b}^2(\boldsymbol{x}_i)\right)\right)\right]$$

$$= \sum_{i=1}^{N_{MCS}}\left(\boldsymbol{\Sigma}_{pb\,i,k}\right) + \sum_{i=1}^{N_{MCS}}\left(\boldsymbol{\Sigma}_{pb\,k,i}\right) - {\sigma_b}^2(\boldsymbol{x}_i), i = 1,2,\ldots N_{MCS} \tag{52}$$

Since the matrix is symmetric, the equation above can be further reduced to:

$$\boldsymbol{x}_{tr}^{*} = \arg\max_{x_i \in S} \left\{ 2\sum_{k=1}^{N_{MCS}} \left(\Sigma_{b\,i,k}\right) - \sigma_b{}^2(\boldsymbol{x}_i) \right\}$$
$$= \arg\max_{x_i \in S} \left\{ 2\sum_{k=1}^{N_{MCS}} \left(\rho_{i,j}^{b}\sigma_b(\boldsymbol{x}_i)\sigma_b(\boldsymbol{x}_k)\right) - \sigma_b{}^2(\boldsymbol{x}_i) \right\}, i = 1,2, \dots N_{MCS} \quad (53)$$

This learning strategy takes the Kriging correlation into consideration and the value of this case equals to $v(\boldsymbol{x}_i) = 2\sum_{k=1}^{N_{MCS}} \left(\rho_{i,k}^{b}\sigma_b(\boldsymbol{x}_i)\sigma_b(\boldsymbol{x}_k)\right) - \sigma_b{}^2(\boldsymbol{x}_i)$. Essentially, this learning strategy is prone to selecting the point with large variance and high probability density.

***Lemma 2***. Considering theorem 3 and $\mathrm{N}_{\rho^b<0} \ll \mathrm{N}_{\rho^b\geq 0}$, the following equations always hold:

$$\Gamma(\boldsymbol{x}_i) \propto \left(\sigma_b(\boldsymbol{x}_i)\middle|\rho_{i,k}^{b}\right) \quad (54)$$

And

$$\Gamma(\boldsymbol{x}_i) \propto \left(f(\boldsymbol{x}_i)\middle|\sigma_b(\boldsymbol{x}_i)\right) \quad (55)$$

where $\mathrm{N}_{\rho^b<0}$ and $\mathrm{N}_{\rho^b\geq 0}$ denote the number of elements of the correlation matrix that smaller and greater than zero, respectively, $(A|\mathrm{B})$ denotes A conditional on B (i.e., B is fixed), $f(\boldsymbol{x}_i)$ means the pdf of $\boldsymbol{x}_i$, $\Gamma(\boldsymbol{x}_i)$ can be expressed as follows,

$$\Gamma(\boldsymbol{x}_i) = 2\sum_{k=1}^{N_{MCS}} \left(\rho_{i,k}^{b}\sigma_b(\boldsymbol{x}_i)\sigma_b(\boldsymbol{x}_k)\right) - \sigma_b{}^2(\boldsymbol{x}_i) \quad (56)$$

Where $\sigma_b(\boldsymbol{x}) = \Phi\left(\frac{-\mu_{\hat{g}}(x)}{\sigma_{\hat{g}}(x)}\right) \cdot \Phi\left(\frac{\mu_{\hat{g}}(x)}{\sigma_{\hat{g}}(x)}\right)$ and $\rho_{i,k}^{b}$ can be calculated using Eq. (40).

**Proof.** Taking the derivative of Eq. (56) one can get,

$$\frac{\partial\Gamma(\boldsymbol{x}_i)}{\partial\sigma_b(\boldsymbol{x}_i)} = 2\sum_{k=1}^{N_{MCS}} \left(\rho_{i,k}^{b}\sigma_b(\boldsymbol{x}_k)\right) - 2\sigma_b(\boldsymbol{x}_i)$$
$$= 2\sum_{k=1,\mathrm{k}\neq\mathrm{i}}^{N_{MCS}} \left(\rho_{i,k}^{b}\sigma_b(\boldsymbol{x}_k)\right) + 2\rho_{i,i}^{b}\sigma_b(\boldsymbol{x}_i) - 2\sigma_b(\boldsymbol{x}_i) = 2\sum_{k=1,\mathrm{k}\neq\mathrm{i}}^{N_{MCS}} \left(\rho_{i,k}^{b}\sigma_b(\boldsymbol{x}_k)\right) \geq 0 \quad (57)$$

Therefore,

$$\Gamma(\boldsymbol{x}_i) \propto \left(\sigma_b(\boldsymbol{x}_i)\middle|\rho_{i,k}^{b}\right) \quad (58)$$

Let $\boldsymbol{x}_j$ denote the point that satisfies $f(\boldsymbol{x}_j) > f(\boldsymbol{x}_i)$ and $\sigma_b{}^2(\boldsymbol{x}_i) = \sigma_b{}^2(\boldsymbol{x}_j)$. Thus,

$$\Gamma(\boldsymbol{x}_j) - \Gamma(\boldsymbol{x}_i) = 2\sum_{k=1}^{N_{MCS}} \left(\rho_{j,k}^b \sigma_b(\boldsymbol{x}_j)\sigma_b(\boldsymbol{x}_k)\right) - \sigma_b{}^2(\boldsymbol{x}_j) - \left(2\sum_{k=1}^{N_{MCS}} \left(\rho_{i,k}^b \sigma_b(\boldsymbol{x}_i)\sigma_b(\boldsymbol{x}_k)\right) - \sigma_b{}^2(\boldsymbol{x}_i)\right)$$

$$= 2\sum_{k=1}^{N_{MCS}} \left(\rho_{j,k}^b \sigma_b(\boldsymbol{x}_j)\sigma_b(\boldsymbol{x}_k)\right) - 2\sum_{k=1}^{N_{MCS}} \left(\rho_{i,k}^b \sigma_b(\boldsymbol{x}_i)\sigma_b(\boldsymbol{x}_k)\right)$$

$$= 2\sigma_b(\boldsymbol{x}_i)\sum_{k=1}^{N_{MCS}} (\rho_{j,k}^b - \rho_{i,k}^b)\sigma_b(\boldsymbol{x}_k) \quad (59)$$

Given that $\rho_{i,k}^b \propto \rho_\rho$ and $f(\boldsymbol{x}_j) > f(\boldsymbol{x}_i)$, one can infer:

$$\sum_{k=1}^{N_{MCS}} (\rho_{j,k}^b - \rho_{i,k}^b)\sigma_b(\boldsymbol{x}_k) \geq 0 \quad (60)$$

Thus,

$$\Gamma(\boldsymbol{x}_j) - \Gamma(\boldsymbol{x}_i) \geq 0 \quad (61)$$

## 5. Multiple samples enrichment

The optimal learning strategies discussed above are based on the computational environment that only one computational facility is available. However, it fails to satisfy the computational requirement for multiple training points enrichment. Therefore, the scheme that enriches multiple training samples should be deeply explored in order to satisfy this requirement. Note that the variance of the stochastic estimator(i.e., $\mathrm{Var}\left[\tilde{P}_f^{mc}\right]$ or $\mathrm{Var}\left[\tilde{P}_f^{mi}\right]$) is updated if new training samples are enriched. Note that both the two optimal learning strategies with/without considering Kriging correlation rely on the computation of covariance matrix and the covariance matrix $\boldsymbol{\Sigma}_b$ can be updated after the new training sample is enriched. Concerning this point, a sequential approach for multiple training samples is explored. Let $\boldsymbol{\Sigma}_b^{\{n_{tr}\}}$ denote the covariance matrix after the order of the $n_{tr}$ training sample is enriched. Therefore, the order of the $n_{tr}+1$ training sample is explored according to $\boldsymbol{\Sigma}_b^{\{n_{tr}\}}$. However, the true value of $\boldsymbol{\Sigma}_b^{\{n_{tr}\}}$ is usually unavailable in the process of sequential enrichment of multiple training samples, a well-defined matrix $\widetilde{\boldsymbol{\Sigma}}_b^{\{n_{tr}\}}$ should take place of the true matrix $\boldsymbol{\Sigma}_b^{\{n_{tr}\}}$. Toward this goal, let $\mathrm{L}\left(\boldsymbol{\Sigma}_{i,j}^{\{n_{tr}\}}, \widetilde{\boldsymbol{\Sigma}}_{i,j}^{\{n_{tr}\}}\right)$ be the loss function representing the difference between the true and estimated elements of $\boldsymbol{\Sigma}_b^{\{n_{tr}\}}$ and $\widetilde{\boldsymbol{\Sigma}}_b^{\{n_{tr}\}}$ after the $n_{tr}$ training sample is enriched. The Bayesian estimate of $\widetilde{\boldsymbol{\Sigma}}_{i,j}^{\{n_{tr}\}}$ can be therefore represented as:

$$\widetilde{\boldsymbol{\Sigma}}_{i.j}^{\{n_{tr}\}} = \underset{\boldsymbol{M}_{i,j} \in \mathbb{R}}{\arg\min}\ \mathrm{E}\left[\mathrm{L}\left(\boldsymbol{\Sigma}_{i,j}^{\{n_{tr}\}}, M_{i,j}\right)\right] \quad (62)$$

Where $M_{i,j}$ denotes the element of matrix $\boldsymbol{M}$ in the $i^{th}$ row and $j^{th}$ column. Then the sequential learning with multiple samples strategy can be conducted on $\widetilde{\boldsymbol{\Sigma}}_b^{\{n_{tr}\}}$. To improve the readability, let $\tilde{X}$ denote $\widetilde{\boldsymbol{\Sigma}}_{i.j}^{\{n_{tr}\}}$ and $X$ denote $\boldsymbol{\Sigma}_{i,j}^{\{n_{tr}\}}$, considering the following theorem:

***Theorem 4***. The value of $\tilde{X}$ is equal to the mean, median and mode of $X$, respectively, in association with the loss function defined as **(i)** $L_{MMSE}(\tilde{X}, X) = (\tilde{X} - X)^2$ **(ii)** $L_{MMAE}(\tilde{X}, X) = |\tilde{X} - X|$ and **(iii)** $L_{MAPE}(\tilde{X}, X) = 0$ if $|\tilde{X} - X| \leq C$; and $1$, otherwise, where $0 < \mathrm{C} \ll 1$, respectively.

**Proof**: For the **case (i)**, let $Ł_1 = \mathrm{E}\left[L_{MMSE}(\tilde{X}, X)\right]$:

$$Ł_1 = \mathrm{E}\left[L_{MMSE}(\tilde{X}, X)\right] = \mathrm{E}\left[\left(\tilde{X} - X\right)^2\right]$$
$$= \mathrm{E}[\tilde{X}^2 - 2\tilde{X}X + X^2] = \mathrm{E}[X^2] - 2\tilde{X}\mathrm{E}[X] + \tilde{X}^2 \quad (63)$$

Thus,

$$\frac{\partial Ł_1}{\partial \tilde{X}} = \frac{\partial[\mathrm{E}[X^2] - 2\tilde{X}\mathrm{E}[X] + \tilde{X}^2]}{\partial \tilde{X}} = 2\tilde{X} - 2\mathrm{E}[X] \quad (64)$$

And

$$\frac{\partial^2 Ł_1}{\partial \tilde{X}^2} = \frac{\partial\left[2\tilde{X} - 2\mathrm{E}[X]\right]}{\partial \tilde{X}} = 2 \quad (65)$$

Function $Ł_1$ is convex and exists minima at $\tilde{X} = \mathrm{E}[X]$. This estimation also takes reference to *Minimum Mean Squared Error(MMSE)* estimate [43].

For the **case (ii)**, let $Ł_2 = \mathrm{E}\left[L_{MMAE}(\tilde{X}, X)\right]$ and concerning the fact that $\mathrm{E}[Y] = \int_0^\infty P(Y > y)\, dy$ if random variable $Y$ satisfying $\mathrm{P}(Y \geq 0) = 1$:

$$Ł_2 = \mathrm{E}\left[L_{MMAE}(\tilde{X}, X)\right] = \mathrm{E}\left[\left|\tilde{X} - X\right|\right] = \int_0^\infty P\left(\left|\tilde{X} - X\right| > y\right) dy$$
$$= \int_0^\infty P\left(X > y + \tilde{X}\right) dy + \int_0^\infty P\left(X < \tilde{X} - y\right) dy \quad (66)$$

Let $q_1$ equal to $y + \tilde{X}$ in the first integral and $q_2$ equal to $\tilde{X} - y$ in the second integral, one can get:

$$Ł_2 = \mathrm{E}\left[L_{MMAE}(\tilde{X}, X)\right] = \int_{\tilde{X}}^\infty P(X > q_1) dq_1 + \int_{-\infty}^{\tilde{X}} P(\mathrm{X} < q_2) dq_2 \quad (67)$$

It is differentiable, thus:

$$\frac{\partial Ł_2}{\partial \tilde{X}} = P\left(X < \tilde{X}\right) - P\left(X > \tilde{X}\right) \quad (67)$$

To explore the point that minimizes $Ł_2$, the following equation should be satisfied:

$$P\left(X < \tilde{X}\right) = P\left(X > \tilde{X}\right) \quad (68)$$

Therefore, $\tilde{X}$ is equal to the median of $X$. This estimation also refers to *Minimum Mean Absolute Error (MMAE)* estimate [43].

For the **case (iii)** , let $Ł_3 = \mathrm{E}\left[L_{MAPE}(\tilde{X}, X)\right]$ and the equation can be expressed as follows,

$$Ł_3 = \mathrm{E}\left[L_{MAPE}(\tilde{X}, X)\right] = \mathrm{P}\left(\left|\mathrm{X} - \tilde{X}\right| > C\right) \cdot 1 + \mathrm{P}\left(\left|X - \tilde{X}\right| \leq 0\right) \cdot 0$$
$$= 1 - \mathrm{P}\left(\left|X - \tilde{X}\right| \leq C\right) = 1 - 2\int_{\tilde{X}-C}^{\tilde{X}+C} \rho_X(x) dx = 1 - 2C\rho_{\mathrm{X}}\left(\tilde{X}\right) \quad (69)$$

Where $\rho_X(x)$ denotes the pdf of random variable $X$. To minimize $Ł_3$, the term $2C\rho_X(x)$ should be maximized, which means that $v$ researches the mode point with maximum pdf. This is also known as *Maximum A posteriori Probability Estimate (MAPE)* [43].

According to *theorem 4*, the MMSE of $\widetilde{\mathbf{\Sigma}}_{i,j}^{\{n_{tr}\}}$ is equal to the mean of $\mathbf{\Sigma}_{i,j}^{\{n_{tr}\}}$, which can be represented as:

$$\widetilde{\mathbf{\Sigma}}_{i,j}^{\{n_{tr}\}MMSE} = \mathrm{E}\left[\mathbf{\Sigma}_{i,j}^{\{n_{tr}\}}\right] = \int_{-\infty}^{\infty} \mathbf{\Sigma}_{i,j}^{\{n_{tr}\}}(\boldsymbol{x}_{tr}^{*}, y)\varphi\big(y|\mu_{\hat{g}}(\boldsymbol{x}_{tr}^{*}), \sigma_{\hat{g}}^{2}(\boldsymbol{x}_{tr}^{*})\big)dy \tag{70}$$

where $\mathbf{\Sigma}_{i,j}^{\{n_{tr}\}}(\boldsymbol{x}_{tr}^{*}, y)$ denotes the elements of the covariance matrix $\mathbf{\Sigma}_{b}^{\{n_{tr}\}}$ after the selected training point $\boldsymbol{x}_{tr}^{*}$ with the defined response $y$ enriched and $\varphi\big(y|\mu_{\hat{g}}(\boldsymbol{x}_{tr}^{*}), \sigma_{\hat{g}}^{2}(\boldsymbol{x}_{tr}^{*})\big)$ denotes the Gaussian probability of $y$ with the parameter mean $\mu_{\hat{g}}(\boldsymbol{x}_{tr}^{*})$ and variance $\sigma_{\hat{g}}^{2}(\boldsymbol{x}_{tr}^{*})$. This integration is achieved by the technique of Gaussian integral which is represented in the Appendix A3. And the MMAE of $\widetilde{\mathbf{\Sigma}}_{i,j}^{\{n_{tr}\}}$ is equal to the median of $\mathbf{\Sigma}_{i,j}^{\{n_{tr}\}}$, which means:

$$\widetilde{\mathbf{\Sigma}}_{i,j}^{\{n_{tr}\}MMAE} = \tau_{\mathbf{\Sigma}_{i,j}^{\{n_{tr}\}}}^{-1}\left(\frac{1}{2}\right) \tag{71}$$

where $\tau_{\mathbf{\Sigma}_{i,j}^{\{n_{tr}\}}}^{-1}$ denotes the inverse of the CDF function of $\mathbf{\Sigma}_{i,j}^{\{n_{tr}\}}$. However, the numerical implementation of MMAE is non-trivial compared to the MMSE and MAPE approaches. It requires large number of sampling to estimate the median value. Hence, it is not suitable for algorithm implementation while it can be computationally efficient in terms of the performance of reducing of calls to the limit state function. Moreover, the MAPE of $\widetilde{\mathbf{\Sigma}}_{i,j}^{\{n_{tr}\}}$ is equal to the mode of $\mathbf{\Sigma}_{i,j}^{\{n_{tr}\}}$, which can be represented as:

$$\widetilde{\mathbf{\Sigma}}_{i,j}^{\{n_{tr}\}MAPE} = \mathbf{\Sigma}_{i,j}^{\{n_{tr}\}}\left(\boldsymbol{x}_{tr}^{*}, \mu_{\hat{g}}(\boldsymbol{x}_{tr}^{*})\right) \tag{72}$$

MAPE requires only one point, which is the Gaussian mean of the output from the Kriging model. Among the three estimates, the MAPE approach is the most efficient one that requires least computational time-complexity.

## 6. Numerical Investigation

In this section, two widely studied benchmark problems are explored to investigate the computational performance of learning functions including EFF [17], U [16], LIF [24], H [23], REIF [26] and FNEIF [27], proposed optimal without considering Kriging correlation(denoted as $\mathbb{L}_{opt}^{nco}$) and optimal considering Kriging correlation learning strategies(denoted as $\mathbb{L}_{opt}^{wco}$). Parameter sets mainly including the initial hyper-parameter for kernel functions, number of initial training samples(DoE), candidate design samples pool and so on are set exactly same. Computational efficiency are represented in terms of the average number of calls to the performance function with corresponding coefficient of variation (i.e., $\overline{N}_{call}$ and $COV$ of $N_{call}$), the average number of iterations with corresponding coefficient of variation (i.e., $\overline{N}_{ite}$ and COV of $N_{ite}$) and the average error (i.e., $\epsilon = \left|\hat{P}_f^{MCS} - \hat{P}_f\right| / \hat{P}_f^{MCS}$) with corresponding COV. To keep consistency, the stopping criterion corresponding to the stochastic estimator of failure probability(i.e., $\tilde{P}_f^{mi}$ and $\tilde{P}_f^{mc}$) is set as $\sigma_{\tilde{P}_f} / \mu_{\tilde{P}_f} \leq 10^{-3}$.

### 6.1 Rastrigin function

The first example is a modified Rastrigin function, which has been investigated as a benchmark problem in many literatures [2], [10], [21]. Random variables $x_1$ and $x_2$ for this problem all follow mutually independent standard normal distributions (e.g. mean of 0 and standard deviation of 1). The performance

function can be read as:

$$g(x_1, x_2) = 10 - \sum_{i=1}^{2} \left( x_i^2 - 5cos(2\pi x_i) \right) \tag{73}$$

For this example, the coefficient of variation of failure probability, $COV_{P_f}$, is set to be as small as 0.05, the number of candidate design samples is set as $1 \times 10^4$. Comparative results for various learning functions are reported in Table 1. Results show that $\mathbb{L}_{opt}^{wco}$ only requires 312.35 number of evaluations to the performance function, which is among the least of all the represented learning strategies. For this example, the learning strategy considering Kriging correlation $\mathbb{L}_{opt}^{wco}$ overperforms than $\mathbb{L}_{opt}^{nco}$. Moreover, theorem 1 represented in this article can be demonstrated by the fact that simulation data of U and $\mathbb{L}_{opt}^{wco}$ learning strategy is exactly same. However, the computational performance of $\mathbb{L}_{opt}^{wco}$ is not sufficiently robust in terms of the COV of $N_{call}$ because its COV is among the largest one with 0.1389. According to Table 1, the FNEIF learning function is the most stable one for this example because its COV of $N_{call}$ is as small as 0.027. However, FNEIF learning function needs the average number of evaluations as large as 371.94. Moreover, all the learning strategies can achieve very accurate result in terms of the average relative error $\bar{\epsilon}$ is smaller than 0.01.

**Table 1.** Reliability analysis results of example 1 with EFF, U, H, LIF, REIF, FNEIF, proposed optimal without considering Kriging correlation(denoted as $\mathbb{L}_{opt}^{nco}$) and optimal considering Kriging correlation learning strategies(denoted as $\mathbb{L}_{opt}^{wco}$). $N_{call}$, $N_{ite}$, $\hat{P}_f^{pc}$ and $\epsilon$ denote the number of calls the performance function, the number of iterations, estimated probability of failure and corresponding error. This average performance is based on 100 times simulations.

| Learning strategy | $\bar{N}_{call}$ | COV of $N_{call}$ | $\bar{\epsilon}$ | COV of $\bar{\epsilon}$ |
|---|---|---|---|---|
| MCS | $1 \times 10^4$ | - | - | - |
| EFF | $12 + 333.20$ | 0.0512 | 0.0004 | 0.7404 |
| U | $12 + 338.78$ | 0.1019 | 0.0016 | 1.7689 |
| H | $12 + 402.57$ | 0.0348 | 0.0005 | 0.7321 |
| LIF | $12 + 324.26$ | 0.0361 | 0.0004 | 0.7044 |
| REIF | $12 + 317.58$ | 0.0389 | 0.0003 | 1.0287 |
| FNEIF | $12 + 359.94$ | 0.0270 | 0.0005 | 0.5894 |
| $\mathbb{L}_{opt}^{nco}$ | $12 + 338.78$ | 0.1019 | 0.0016 | 1.7689 |
| $\mathbb{L}_{opt}^{wco}$ | $12 + 300.35$ | 0.1389 | 0.0038 | 1.4835 |

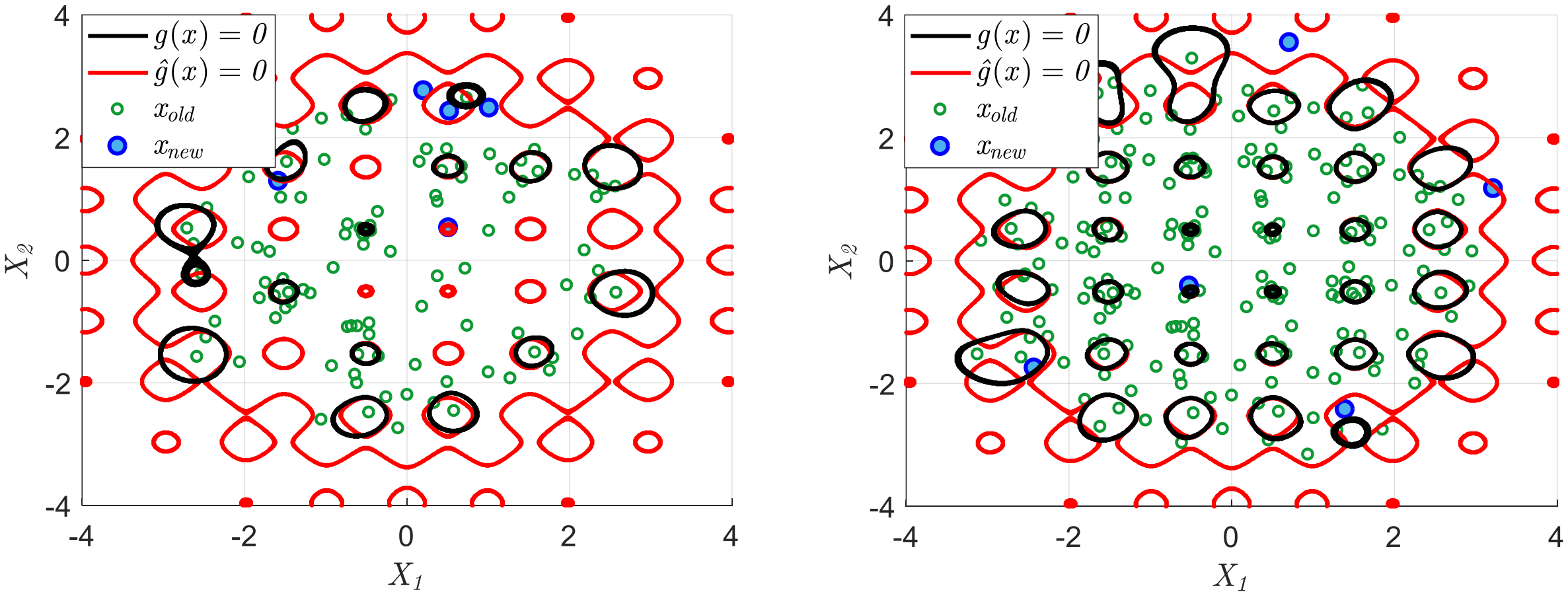

(a) (b)

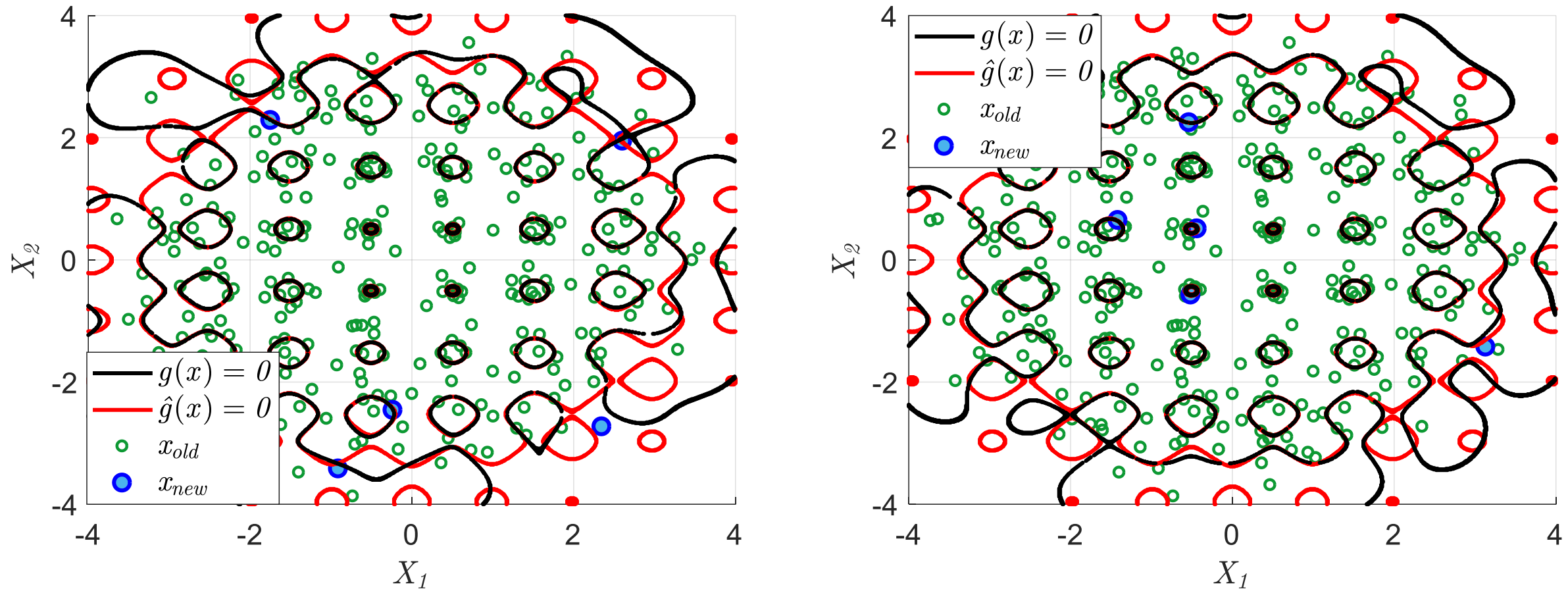

(c) (d)

**Fig. 1**. Evolution of the limit state with $\mathbb{L}_{opt}^{wco}$ enhanced with MMSE ($N_{para} = 5$) based on (a) $N_{call} = 100$ (b) $N_{call} = 200$ (b) $N_{ite}$ (c) $N_{call} = 300$ (d) $N_{call} = 400$

To investigate the computational performance of proposed optimal learning function enhanced by the parallel learning strategy, illustrations of evolution of the limit state with $\mathbb{L}_{opt}^{wco}$ enhanced with MMSE and MAPE based on $N_{para} = 5$ and $N_{call} = 100,\ 200, 300$ and $400$ are schemed in Fig 2 and 3. Illustrative results indicate that new training points $\boldsymbol{x}_{new}$ are all located in vicinity of the limit state and keep distance from each other in different regions before the final training process of Kriging surrogate model. However, they are very close at the early stage according to Fig 1(a) and 2(a). Moreover, the corresponding boxplots of $N_{call}$ vs $N_{para}$ and $N_{ite}$ vs $N_{para}$ with $N_{para} = 3, 6, 9, 12, 15$ and $18$ are illustrated in Fig 3. According to Table 1 and Fig 1 to 3, the number of iteration with parallel learning strategy (MMSE or MAPE) is significantly reduced compared to the approach via single training point-based enrichment. For example, $N_{call}$ for $\mathbb{L}_{opt}^{wco}$ with single point enrichment is 312.35. Nevertheless, $N_{call}$ for MMSE parallel learning strategies with $N_{para} = 3$, $N_{para} = 6$, $N_{para} = 9$ and $N_{para} = 12$ training points-based enrichment only need 104.3, 53.2, 36.7 and 32.5 iterations. According to Fig 1 to 3, both the MMSE and MAPE shows great efficiency in enriching multiple training points because the total number of calls to the performance function are almost unchanged despite $N_{para}$ increases. For example, the average number of calls to the performance, $\bar{N}_{call}$ is 332.35 when $N_{para} = 3$ by MMSE compared to $\bar{N}_{call} = 362.9$ when $N_{para} = 12$. However, the number of iterations reduces significantly. For example, $\bar{N}_{it}$ is 104.3 when $N_{para} = 3$ by MMSE compared to $\bar{N}_{it} = 32.5$ when $N_{para} = 12$. Moreover, MAPE is more efficient compared to MMSE in terms of the computationally algorithmic advantage due to the fact that MAPE only requires one evaluation to the Kriging model but MMSE needs to estimate several times by using the Gaussian integral. Moreover, the speed of the decrement of $\bar{N}_{ite}$ is fast when $N_{para}$ is small($\leq 12$) according to Fig. 3 (c) and (d), which means that sufficient computers can not always guarantee the optimal strategy to reduce the number of iterations. In other words, preparing too much computers for parallelization computing may cause computational waste. For example, $N_{ite}$s are 24.5 and 23.2 for MAPE with $N_{para} = 12$ and 15 training points enrichment as shown in Fig 3(d), which are very close to each other. Hence, an appropriate definition of the number of multiple training point enrichment for parallel learning strategies is necessary, which, however, is not explored in this paper.

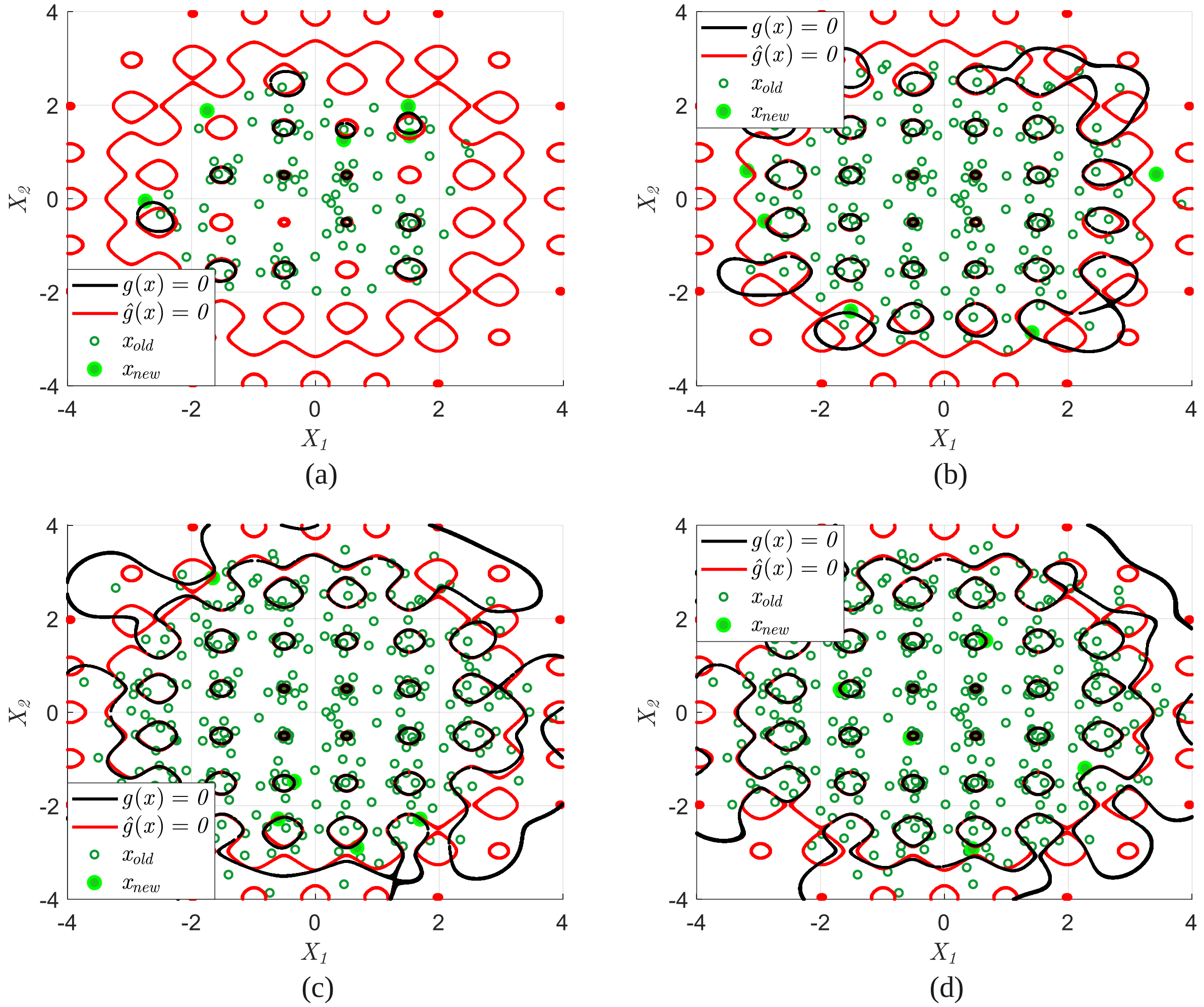

**Fig. 2**. Evolution of the limit state with $\mathbb{L}_{opt}^{wco}$ enhanced with MAPE ($N_{para}$ = 5) based on (a) $N_{call}$ = 100 (b) $N_{call}$ = 200 (b) $N_{ite}$ (c) $N_{call}$ = 300 (d) $N_{call}$ = 400

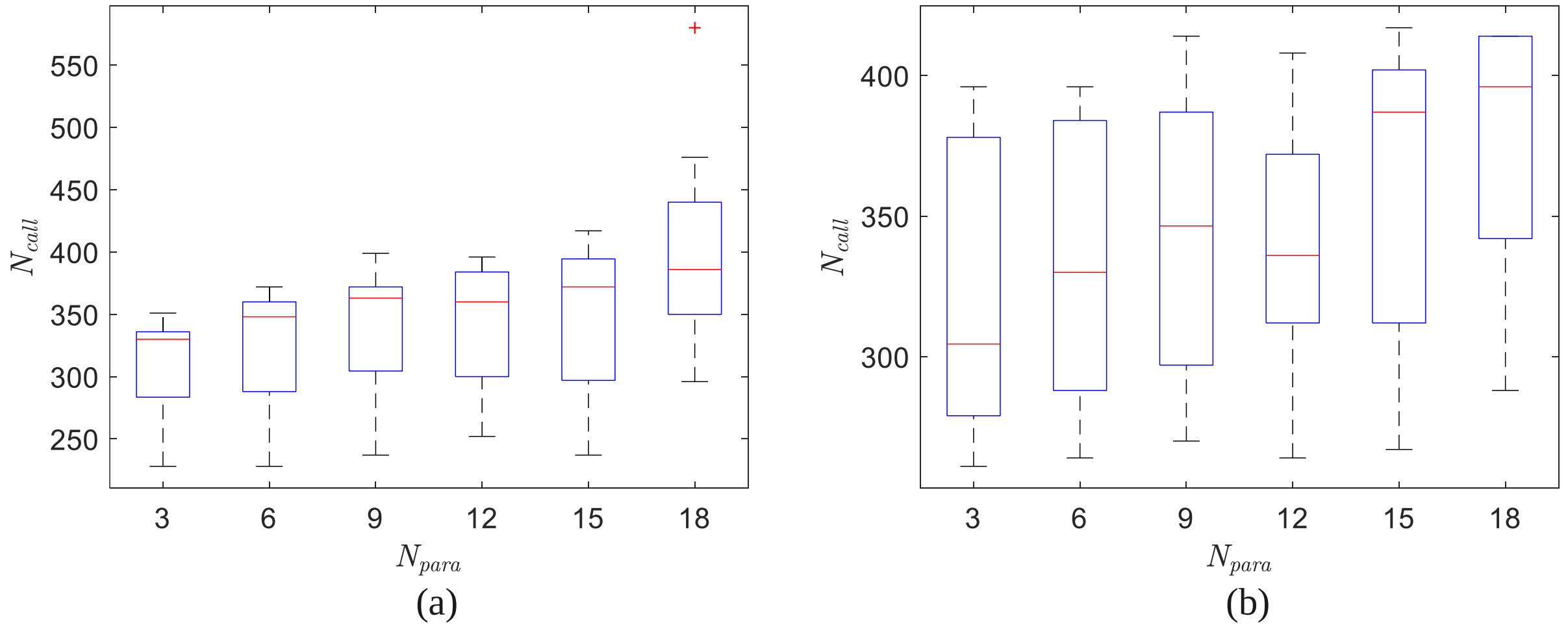

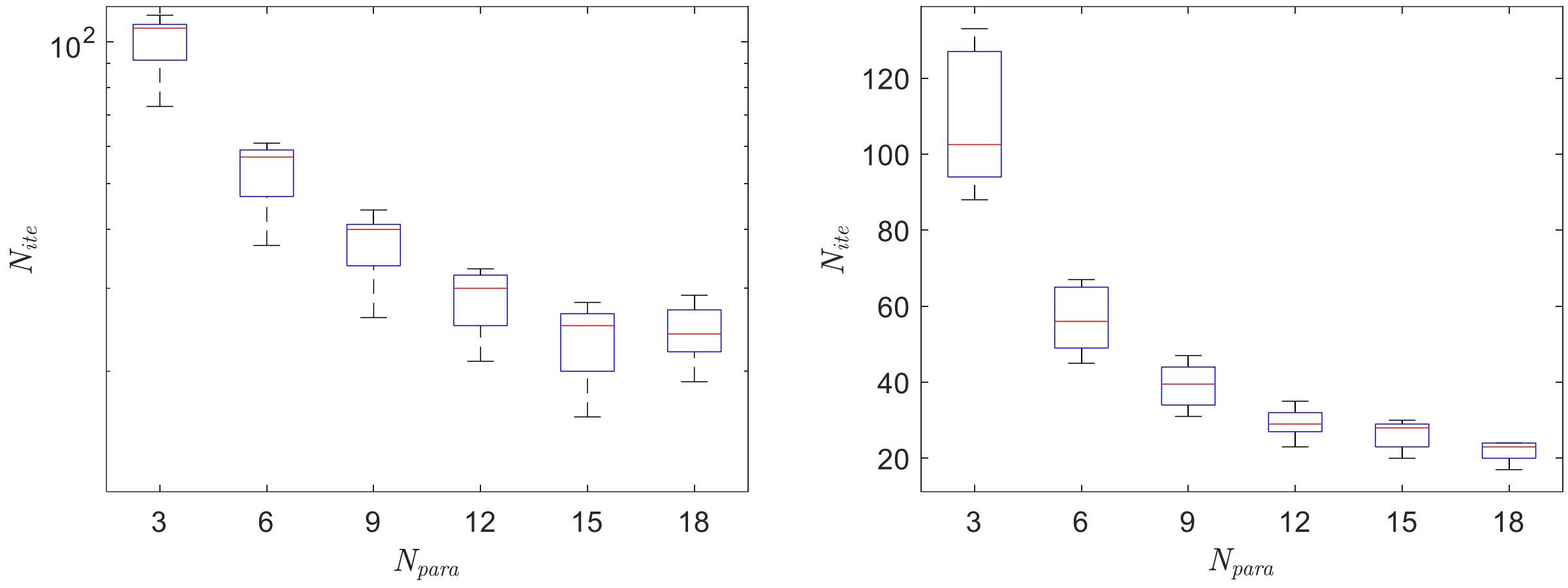

(c)

(d)

**Fig. 3**. Boxplots of $\mathbb{L}_{opt}^{wco}$ enhanced with the parallel learning strategy of (a) $N_{call}$ vs $N_{para}$ with MMSE (b) $N_{ite}$ vs $N_{para}$ with MMSE (c) $N_{ite}$ vs $N_{para}$ with MAPE (d) $N_{ite}$ vs $N_{para}$ with MAPE

## 8. Conclusion

This paper proposes several theorems to explore the theoretical optimal learning strategy for adaptive Kriging-based reliability analysis. Toward this goal, the variance of the estimated probability of failure is first quantified, which facilitates the criteria for defining the optimal learning strategy. The mathematical expressions of the optimal learning strategy with/without considering Kriging correlation are derived based on the proposed criteria for optimal definition. It is interesting that the well-known $U$ learning function belongs to one of the forms of optimal learning function without considering Kriging correlation. However, the implementation for optimal learning strategy that considers Kriging correlation is shown to be intensively computational demanding, which needs to use large computational resource. This is in part due to the requirement for the storage of covariance matrix. Moreover, this paper also provides a way to explore the optimal parallel learning strategy, which is based on the framework of minimizing the loss function. Two widely used numerical examples are implemented to demonstrate our proposed theorems. Moreover, the Gaussian process can be also leveraged to improve the computational efficiency in computational science [43-50], transportation engineering[51-55], biological engineering[56,57] and artificial intelligence[58-65].

## Appendix

### A1. Learning functions from the literatures

In this article, some state-of-the-art learning functions including EFF [17], U [16], LIF [24], H [23], REIF [26] and FNEIF [27] are briefly introduced, computational details can be referred to the literatures. First, by quantifying the probability of making wrong sign estimate (+/-) in $\hat{g}(\boldsymbol{x})$. $U$ learning function is proposed to define this uncertainty U [16]. Thus, the $U$ learning function can be defined as.,

$$U(\boldsymbol{x}) = \frac{|\mu_{\hat{g}}(\boldsymbol{x})|}{\sigma_{\hat{g}}(\boldsymbol{x})} \tag{A.1}$$

The point that minimizes the response of $U$ is selected in the learning iteration. And the stopping criterion for this learning function can be set as $min(U(\boldsymbol{x})) \geq 2$. In $EFF$, the proximity of points to the limit state $g(\boldsymbol{x}) = 0$ and their variance are the two key factors [17]. The mathematical expression of $EFF$ is presented below:

$$\begin{aligned} EFF(\boldsymbol{x}) = \mu_{\hat{g}}(\boldsymbol{x}) &\left[2\Phi\left(\frac{-\mu_{\hat{g}}(\boldsymbol{x})}{\sigma_{\hat{g}}(\boldsymbol{x})}\right) - \Phi\left(\frac{a^- - \mu_{\hat{g}}(\boldsymbol{x})}{\sigma_{\hat{g}}(\boldsymbol{x})}\right) - \Phi\left(\frac{a^+ - \mu_{\hat{g}}(\boldsymbol{x})}{\sigma_{\hat{g}}(\boldsymbol{x})}\right)\right] \\ -\sigma_{\hat{g}}(\boldsymbol{x}) &\left[2\phi\left(\frac{-\mu_{\hat{g}}(\boldsymbol{x})}{\sigma_{\hat{g}}(\boldsymbol{x})}\right) - \phi\left(\frac{a^- - \mu_{\hat{g}}(\boldsymbol{x})}{\sigma_{\hat{g}}(\boldsymbol{x})}\right) - \phi\left(\frac{a^+ - \mu_{\hat{g}}(\boldsymbol{x})}{\sigma_{\hat{g}}(\boldsymbol{x})}\right)\right] \\ +2\sigma_{\hat{g}}(\boldsymbol{x}) &\left[\Phi\left(\frac{a^+ - \mu_{\hat{g}}(\boldsymbol{x})}{\sigma_{\hat{g}}(\boldsymbol{x})}\right) - \Phi\left(\frac{a^- - \mu_{\hat{g}}(\boldsymbol{x})}{\sigma_{\hat{g}}(\boldsymbol{x})}\right)\right] \end{aligned} \tag{A.2}$$

where $\phi(\cdot)$ is the standard normal probability density function, $\delta(\boldsymbol{x}) = 2\sigma_{\hat{g}}(\boldsymbol{x})$, $a^+ = \delta(\boldsymbol{x})$ and $a^- = -\delta(\boldsymbol{x})$. The term $[\delta(\boldsymbol{x}) - |h|]$ in Eq. (A.2) measures if the target point is close to the limit state with the cumulated probability density in the interval $[a^-, a^+]$, which can be reflected in the term $\phi(h; \mu_{\hat{g}}(\boldsymbol{x}), \sigma_{\hat{g}}(\boldsymbol{x}))$.The point that maximizes the $EFF$ response is chosen as the next best training point to refine the Kriging model with the corresponding stopping criterion expressed as $max(EFF(\boldsymbol{x})) \leq 0.001$. Based on the entropy theory, the learning function H can be defined as [23]:

$$\begin{aligned} H(\boldsymbol{x}) = \Bigg|\ln\left(\sqrt{2\pi}\sigma_{\hat{g}}(\boldsymbol{x}) + \frac{1}{2}\right)&\left[\Phi\left(\frac{2\sigma_{\hat{g}}(\boldsymbol{x}) - \mu_{\hat{g}}(\boldsymbol{x})}{\sigma_{\hat{g}}(\boldsymbol{x})}\right) - \Phi\left(\frac{-2\sigma_{\hat{g}}(\boldsymbol{x}) - \mu_{\hat{g}}(\boldsymbol{x})}{\sigma_{\hat{g}}(\boldsymbol{x})}\right)\right] \\ &- \left[\frac{2\sigma_{\hat{g}}(\boldsymbol{x}) - \mu_{\hat{g}}(\boldsymbol{x})}{2}\phi\left(\frac{2\sigma_{\hat{g}}(\boldsymbol{x}) - \mu_{\hat{g}}(\boldsymbol{x})}{\sigma_{\hat{g}}(\boldsymbol{x})}\right) \right. \\ &\left. + \frac{2\sigma_{\hat{g}}(\boldsymbol{x}) + \mu_{\hat{g}}(\boldsymbol{x})}{2}\phi\left(\frac{2\sigma_{\hat{g}}(\boldsymbol{x}) - \mu_{\hat{g}}(\boldsymbol{x})}{\sigma_{\hat{g}}(\boldsymbol{x})}\right)\right]\Bigg| \end{aligned} \tag{A.3}$$

The next best training point is selected as the one that maximizes the value of H. By taking the probability density into consideration, the LIF learning can be expressed as below [24]:

$$LIF(\boldsymbol{x}) = \begin{cases} \Phi\left(-U(\boldsymbol{x})f(\boldsymbol{x})\left[\mu_{\hat{g}}^{N}(\boldsymbol{x}) + \sum_{m=1}^{N/2} C_n^{2m}\mu_{\hat{g}}^{N-2m}(\boldsymbol{x})\,\sigma_{\hat{g}}^{2m}(\boldsymbol{x})(2m-1)!\right]\right), if\ N\ is\ even \\ \Phi\left(-U(\boldsymbol{x})f(\boldsymbol{x})\left[\sqrt{\frac{2}{\pi}}\sum_{m=0}^{N} C_N^{m}\mu_{\hat{g}}^{N-m}(\boldsymbol{x})\,\sigma_{\hat{g}}^{m}(\boldsymbol{x})\int_{-\frac{\mu_{\hat{g}}(x)}{\sigma_{\hat{g}}(x)}}^{+\infty} t^m e^{-\frac{t^2}{2}}dt\right]\right), if\ N\ is\ odd \end{cases} \tag{A.4}$$

where $f(\boldsymbol{x})$ denotes the joint pdf of variable $\boldsymbol{x}$ and $N$ is the corresponding dimension number. Moreover, $C_x^y = \frac{x!}{y!(x-y)!}$ denotes the operator of combination. LIF also follows the rule that the point is selected as the next point if it maximizes the value of LIF. In addition, the REIF and REIF2 learning functions are also estimated as:

$$\begin{aligned} REIF(\boldsymbol{x}) &= c_\omega\sigma_{\hat{g}}(\boldsymbol{x}) - \mu_f(\boldsymbol{x}) \\ REIF2(\boldsymbol{x}) &= \left[c_\omega\sigma_{\hat{g}}(\boldsymbol{x}) - \mu_f(\boldsymbol{x})\right]f(\boldsymbol{x}) \end{aligned} \tag{A.5}$$

Where $c_\omega = 2$ is a constant and $\mu_f(\boldsymbol{x})$ can be estimated as:

$$\mu_f(\boldsymbol{x}) = \sqrt{\frac{2}{\pi}}\sigma_{\hat{g}}(\boldsymbol{x})\exp\left(-\frac{\mu_{\hat{g}}^2(\boldsymbol{x})}{2\sigma_{\hat{g}}^2(\boldsymbol{x})}\right) + \mu_{\hat{g}}(\boldsymbol{x})\left[2\Phi\left(\frac{\mu_{\hat{g}}(\boldsymbol{x})}{\sigma_{\hat{g}}(\boldsymbol{x})}\right) - 1\right] \tag{A.6}$$

Candidate design points that maximize $REIF(\boldsymbol{x})$ and $REIF2(\boldsymbol{x})$ are selected next best training points. Following this work, the FNEIF learning function is also proposed by Shi et al., [27]. The mathematical expression of FNEIF can be represented as follows [27]. First, for $2\sigma_f(\boldsymbol{x}) \geq \mu_f(\boldsymbol{x})$:

$$\begin{aligned} FNEIF(\boldsymbol{x}) = {} & 2\sigma_f(\boldsymbol{x})\left[\Phi\left(\frac{2\sigma_f(\boldsymbol{x}) - \mu_{\hat{g}}(\boldsymbol{x})}{\sigma_{\hat{g}}(\boldsymbol{x})}\right) - \Phi\left(\frac{-2\sigma_f(\boldsymbol{x}) - \mu_{\hat{g}}(\boldsymbol{x})}{\sigma_{\hat{g}}(\boldsymbol{x})}\right)\right] + \\ & \mu_f(\boldsymbol{x})\left[\begin{array}{c} \Phi\left(\frac{\mu_f(\boldsymbol{x}) - \mu_{\hat{g}}(\boldsymbol{x})}{\sigma_{\hat{g}}(\boldsymbol{x})}\right) - \Phi\left(\frac{-\mu_f(\boldsymbol{x}) - \mu_{\hat{g}}(\boldsymbol{x})}{\sigma_{\hat{g}}(\boldsymbol{x})}\right) - \\ \Phi\left(\frac{2\sigma_f(\boldsymbol{x}) - \mu_{\hat{g}}(\boldsymbol{x})}{\sigma_{\hat{g}}(\boldsymbol{x})}\right) + \Phi\left(\frac{-2\sigma_f(\boldsymbol{x}) - \mu_{\hat{g}}(\boldsymbol{x})}{\sigma_{\hat{g}}(\boldsymbol{x})}\right) - 1 \end{array}\right] + \\ & \sigma_{\hat{g}}(\boldsymbol{x})\left[\phi\left(\frac{\mu_f(\boldsymbol{x}) + \mu_{\hat{g}}(\boldsymbol{x})}{\sigma_{\hat{g}}(\boldsymbol{x})}\right) + \phi\left(\frac{\mu_f(\boldsymbol{x}) - \mu_{\hat{g}}(\boldsymbol{x})}{\sigma_{\hat{g}}(\boldsymbol{x})}\right)\right] + \\ & \mu_{\hat{g}}(\boldsymbol{x})\left[\Phi\left(\frac{\mu_f(\boldsymbol{x}) + \mu_{\hat{g}}(\boldsymbol{x})}{\sigma_{\hat{g}}(\boldsymbol{x})}\right) - \Phi\left(\frac{\mu_f(\boldsymbol{x}) - \mu_{\hat{g}}(\boldsymbol{x})}{\sigma_{\hat{g}}(\boldsymbol{x})}\right)\right] \end{aligned} \tag{A.7a}$$

And for $2\sigma_f(\boldsymbol{x}) < \mu_f(\boldsymbol{x})$

$$FNEIF(\boldsymbol{x}) = 2\sigma_f(\boldsymbol{x})\left[\Phi\left(\frac{2\sigma_f(\boldsymbol{x}) - \mu_{\hat{g}}(\boldsymbol{x})}{\sigma_{\hat{g}}(\boldsymbol{x})}\right) - \Phi\left(\frac{-2\sigma_f(\boldsymbol{x}) - \mu_{\hat{g}}(\boldsymbol{x})}{\sigma_{\hat{g}}(\boldsymbol{x})}\right)\right] - \mu_f(\boldsymbol{x})$$

$$\sigma_{\hat{g}}(\boldsymbol{x})\left[\phi\left(\frac{2\sigma_f(\boldsymbol{x}) + \mu_{\hat{g}}(\boldsymbol{x})}{\sigma_{\hat{g}}(\boldsymbol{x})}\right) + \phi\left(\frac{2\sigma_f(\boldsymbol{x}) - \mu_{\hat{g}}(\boldsymbol{x})}{\sigma_{\hat{g}}(\boldsymbol{x})}\right)\right] +$$
$$\mu_{\hat{g}}(\boldsymbol{x})\left[\Phi\left(\frac{2\sigma_f(\boldsymbol{x}) + \mu_{\hat{g}}(\boldsymbol{x})}{\sigma_{\hat{g}}(\boldsymbol{x})}\right) - \Phi\left(\frac{2\sigma_f(\boldsymbol{x}) - \mu_{\hat{g}}(\boldsymbol{x})}{\sigma_{\hat{g}}(\boldsymbol{x})}\right)\right] \quad \text{(A.7b)}$$

where $\sigma_f(\boldsymbol{x})$ can be estimated as:

$$\sigma_f(\boldsymbol{x}) = \mu_{\hat{g}}^2(\boldsymbol{x}) + \sigma_{\hat{g}}^2(\boldsymbol{x}) - \mu_f^2(\boldsymbol{x}) \quad \text{(A.8)}$$

The next best training point is selected that maximized the value of FNEIF.

**A2. Computational aspects for the Bivariate Normal Integral**

Computational obstacle regarding Eq. (40) involves the bivariate normal integral. Many computational platforms provide a single computation for bivariate normal integral, however, few of them can achieve parallel computation. To enable parallel computational process for Eq. (40), a way of approximation for the computation of the bivariate normal integral is necessary. According to [24], [69], the approximation of the value of the CDF of bivariate normal distribution can be expressed as:

$$\Phi\left([0,0]; [\mu_i; \mu_j], \begin{bmatrix} \Sigma_{i,i} & \Sigma_{j,i} \\ \Sigma_{i,j} & \Sigma_{j,j} \end{bmatrix}\right) = \Theta(m_i, m_j; \rho_{i,j}) + \Phi_n(m_i) + \Phi_n(m_j) - 1 \quad \text{(A.9)}$$

where $\Phi_n$ denotes the CDF of standard normal distribution, $m_i = -\frac{\mu_i}{\sqrt{\Sigma_{i,i}}}$ and $m_j = -\frac{\mu_j}{\sqrt{\Sigma_{j,j}}}$ are the normalized mean in the standard normal space and $\Theta(\cdot)$ is a function that can be expressed as follows:

$$\Theta(m_i, m_j; \rho_{i,j}) = \frac{1}{2\pi}\int_{cos^{-1}\rho_{i,j}}^{\pi} exp\left\{-\frac{m_i^2 + m_j^2 - 2m_i m_j cosx}{2sin^2 x}\right\}dx \quad \text{(A.10)}$$

And the derivative of $\Theta$ corresponding to $\rho_{i,j}$ can be computed as:

$$\frac{\partial\Theta}{\partial\rho_{i,j}}(m_i, m_j; \rho_{i,j}) = \frac{1}{2\pi\sqrt{1 - \rho_{i,j}^2}} exp\left\{-\frac{m_i^2 + m_j^2 - 2\rho_{i,j} m_i m_j}{2(1 - \rho_{i,j}^2)}\right\} \quad \text{(A.11)}$$

If the computation is approximated by 2 points, $\Theta(m_i, m_j; \rho_{i,j})$ can be approximated as:

$$\Theta(m_i, m_j; \rho_{i,j}) \approx \frac{\rho_{i,j}}{2}\left\{\frac{\partial\Theta}{\partial\rho_{i,j}}\left(m_i, m_j; \frac{3 - \sqrt{3}}{6}\rho_{i,j}\right) + \frac{\partial\Theta}{\partial\rho_{i,j}}\left(m_i, m_j; \frac{3 + \sqrt{3}}{6}\rho_{i,j}\right)\right\}$$
$$+\Phi_n(-m_i)\Phi_n(-m_j) \quad \text{(A.12)}$$

If 3 approximating points are selected, $\Theta(m_i, m_j; \rho_{i,j})$ can be approximated as:

$$\Theta(m_i, m_j; \rho_{i,j}) \approx \frac{\rho_{i,j}}{18}\left\{\begin{matrix} 5\frac{\partial\Theta}{\partial\rho_{i,j}}\left(m_i, m_j; \frac{1-\sqrt{\frac{3}{5}}}{2}\rho_{i,j}\right) + 8\frac{\partial\Theta}{\partial\rho_{i,j}}\left(m_i, m_j; \frac{1}{2}\rho_{i,j}\right) \\ +5\frac{\partial\Theta}{\partial\rho_{i,j}}\left(m_i, m_j; \frac{1+\sqrt{\frac{3}{5}}}{2}\rho_{i,j}\right) \end{matrix}\right\} + \Phi_n(-m_i)\Phi_n(-m_j) \quad (A.13)$$

Other approximate equations with different numbers are available in the literature. Generally, the accuracy of the approximation increase as the number of the points increases. This article uses 2 point-based approximation.

### A3. Computational aspects for Gaussian integral

To improve the computational efficiency and reduce the time-complexity, the Gaussian integral is adopted to estimate the integral in Eq. (70),

$$\widetilde{\boldsymbol{\Sigma}}_{i,j}^{\{n_{tr}\}MMSE} = \mathrm{E}\left[\boldsymbol{\Sigma}_{i,j}^{\{n_{tr}\}}\right] = \int_{-\infty}^{\infty} \boldsymbol{\Sigma}_{i,j}^{\{n_{tr}\}}(\boldsymbol{x}_{tr}^*, y)\varphi\left(y|\mu_{\hat{g}}(\boldsymbol{x}_{tr}^*), \sigma_{\hat{g}}^2(\boldsymbol{x}_{tr}^*)\right)dy \quad (A.14)$$

The integral of Eq. (70) starts from minus infinity and ends in positive infinity. Let $f_{\boldsymbol{\Sigma}}(y) = \boldsymbol{\Sigma}_{i,j}^{\{n_{tr}\}}(\boldsymbol{x}_{tr}^*, y)\varphi\left(y|\mu_{\hat{g}}(\boldsymbol{x}_{tr}^*), \sigma_{\hat{g}}^2(\boldsymbol{x}_{tr}^*)\right)$, and note that it is sufficiently accurate to take place of the interval with $\left[\mu_{\hat{g}}(\boldsymbol{x}_{tr}^*) - 3\sigma_{\hat{g}}(\boldsymbol{x}_{tr}^*),\ \mu_{\hat{g}}(\boldsymbol{x}_{tr}^*) + 3\sigma_{\hat{g}}(\boldsymbol{x}_{tr}^*)\right]$, thus the Eq. (70) can be represented as below:

$$\widetilde{\boldsymbol{\Sigma}}_{i,j}^{\{n_{tr}\}MMSE} = \int_{-\infty}^{\infty} f_{\boldsymbol{\Sigma}}(y)dy \approx \int_{\mu_{\hat{g}}(\boldsymbol{x}_{tr}^*)-3\sigma_{\hat{g}}(\boldsymbol{x}_{tr}^*)}^{\mu_{\hat{g}}(\boldsymbol{x}_{tr}^*)+3\sigma_{\hat{g}}(\boldsymbol{x}_{tr}^*)} f_{\boldsymbol{\Sigma}}(y)dy \quad (A.15)$$

According to the Gaussian integral,

$$\widetilde{\boldsymbol{\Sigma}}_{i,j}^{\{n_{tr}\}MMSE} \approx \int_{\mu_{\hat{g}}(\boldsymbol{x}_{tr}^*)-3\sigma_{\hat{g}}(\boldsymbol{x}_{tr}^*)}^{\mu_{\hat{g}}(\boldsymbol{x}_{tr}^*)+3\sigma_{\hat{g}}(\boldsymbol{x}_{tr}^*)} f_{\boldsymbol{\Sigma}}(y)dy = f_{\boldsymbol{\Sigma}}(y_1)w_1 + f_{\boldsymbol{\Sigma}}(y_2)w_2 \ldots + f_{\boldsymbol{\Sigma}}(y_n)w_n \quad (A.16)$$

where $y_n$ denotes the Gaussian integral position, $w_n$ is the corresponding coefficient and $n$ denotes the number of Gaussian integral points. In this article, three Gaussian integral points are selected (i.e., n = 3), which means that:

$$y_1 = \mu_{\hat{g}}(\boldsymbol{x}_{tr}^*) + 3t_1\sigma_{\hat{g}}(\boldsymbol{x}_{tr}^*),\quad y_2 = \mu_{\hat{g}}(\boldsymbol{x}_{tr}^*) + 3t_2\sigma_{\hat{g}}(\boldsymbol{x}_{tr}^*),\quad y_3 = \mu_{\hat{g}}(\boldsymbol{x}_{tr}^*) + 3t_3\sigma_{\hat{g}}(\boldsymbol{x}_{tr}^*) \quad (A.17)$$

Where $t_1 = 0.7746$, $t_2 = 0$ and $t_3 = -0.7746$ with the corresponding weights $w_1 = 0.5556$, $w_2 = 0.8889$ and $w_3 = -0.5556$.